# Parallel Bayesian Optimization Using Satisficing Thompson Sampling for Time-Sensitive Black-Box Optimization

Xiaobin Song [1], Benben Jiang[1,2]†

† Corresponding author: bbjiang@tsinghua.edu.cn

1 Department of Automation, Tsinghua University, Beijing 100084, China

2 Beijing National Research Center for Information Science and Technology,

Tsinghua University, Beijing 100084, China

**Abstract:** Bayesian optimization (BO) is widely used for black-box optimization problems, and have been shown to perform well in various real-world tasks. However, most of the existing BO methods aim to learn the optimal solution, which may become infeasible when the parameter space is extremely large or the problem is time-sensitive. In these contexts, switching to a satisficing solution that requires less information can result in better performance. In this work, we focus on time-sensitive black-box optimization problems and propose satisficing Thompson sampling-based parallel Bayesian optimization (STS-PBO) approaches, including synchronous and asynchronous versions. We shift the target from an optimal solution to a satisficing solution that is easier to learn. The rate-distortion theory is introduced to construct a loss function that balances the amount of information that needs to be learned with sub-optimality, and the Blahut-Arimoto algorithm is adopted to compute the target solution that reaches the minimum information rate under the distortion limit at each step. Both discounted and undiscounted Bayesian cumulative regret bounds are theoretically derived for the proposed STS-PBO approaches. The effectiveness of the proposed methods is demonstrated on a fast-charging design problem of Lithium-ion batteries. The results are accordant with theoretical analyses, and show that our STS-PBO methods outperform both sequential counterparts and parallel BO with traditional Thompson sampling in both synchronous and asynchronous settings.

## 1. Introduction

Black-box optimization is widely utilized in numerous practical applications, where the objective is to optimize an unknown function $f: \mathcal{X} \to \mathbb{R}$ based on noisy observations [1-3, 31]. For example, tuning hyper-parameters in machine learning algorithms can be considered as such a problem where $x \in \mathcal{X}$ represents a specific set of hyper-parameters and $f(x)$ means the cross-validation accuracy [1]. Similarly, the fast-charging design problem for Lithium-ion batteries can be associated with a black-box optimization task, where implementing a charging protocol $x \in \mathcal{X}$ requires evaluating performance metrics like charging time or cycle life, denoted by $f(x)$ [2]. Given that evaluations for these problems are often resource-intensive and time-consuming, the objective is to minimize the number of evaluations during optimization. A popular technique for addressing such issues is Bayesian optimization (BO), which makes Bayesian assumptions on the unknown function $f(x)$. The BO framework comprises of two main components: (i) a probabilistic surrogate model that approximates the objective function $f(x)$ and (ii) an acquisition function that probes the parameter space $\mathcal{X}$ by balancing between exploration and exploitation. Empirical evidence has showed the effectiveness of BO in diverse applications such as hyper-parameters tuning [1], rapid battery charging design [2], tissue engineering [3], robot path planning [4], and industrial controller design [5, 6].

It is noteworthy that most BO methods aim to learn the optimal solution that either maximizes or minimizes the function $f$. However, when the parameter space $\mathcal{X}$ is extremely large, or when time and resource constraints restrict the number of evaluations, identifying the optimal solution becomes considerably difficult [7]. While in many real-world problems, time sensitivity is an important issue since a complete evaluation may take a considerable amount of time. For example, in fast charging design for Lithium-ion batteries, since batteries are designed to work for a long lifetime, it usually takes months or even years to fully test the cycle life of a battery under a specified protocol [8]. Moreover, the size of the parameter space is much larger than the number of evaluations allowed by the practical problem due to time and resource limitations. In these problems, identifying the optimal solution may become infeasible, and switching to a satisficing solution that can be found in the shortest possible time may perform better. Most BO methods fail in this scenario because they focus on long-term performance and tend to select unexplored points in pursuit of the optimal solution in each iteration. This strategy can become markedly inefficient, particularly as the size of the parameter space expands [7].

Notably, it helps to shift the target from an optimal solution to a satisficing solution for time-sensitive optimization problems [9]. We can appropriately relax the requirement for optimality to reduce the complexity of exploration, given that learning a satisficing solution usually requires less information than the optimal solution, which plays an important role in time-sensitive problems. Therefore, our work focuses on the BO methods of satisficing strategies considering time sensitivity.

Satisficing strategies have been studied in time-sensitive bandit learning [7,9, 16], mainly for Thompson sampling (TS) [10]. Russo and Van Roy [11] analyze TS through information theory and provide a regret bound that depends on the entropy of the optimal-action distribution. It indicates that learning the optimal action becomes extremely difficult when the environment is complex. Therefore, Russo and Van Roy [9] propose satisficing TS that learns a near-optimal action with the requirement of less information. Furthermore, Arumugam and Van Roy [7] develop the Blahut-Arimoto satisficing Thompson sampling (BLASTS) algorithm. They introduce the rate-distortion technique [12, 13] to sequential decision-making problems and propose a rate-distortion function that represents the trade-off between the required information for identifying a satisficing action and its regret. Blahut-Arimoto algorithm [14, 15] is adopted to solve the rate-distortion function. Compared to the satisficing method proposed by Russo and Van Roy [9], the satisficing actions of BLASTS can be computed by the agent, achieving the limit of the rate-distortion function at each time step. Nevertheless, they validate their algorithm's performance using independent multi-armed bandit problems and select linear hypermodels to represent the agent's understanding of the environment. Their approach does not consider the correlation between different actions. In many real-world scenarios, actions are interrelated. For example, in fast-charging design problems, there's a strong correlation among different charging strategies. Taking such interrelationships into account can enhance the efficiency of exploration.

In this work, we propose a satisficing Thompson sampling-based parallel Bayesian optimization (STS-PBO) approach for time-sensitive optimization problems. A surrogate model of Gaussian process is utilized in the proposed method to capture the correlation information among actions. The satisficing Thompson sampling strategy is introduced to reduce the amount of information that needs to be learned. We set the target to an easier-to-learn satisficing solution rather than an optimal solution so that we can identify the target solution within less time, which is helpful for addressing time-sensitive problems. Rate-distortion theory is adopted in this work to formulate the trade-off between sub-optimality and the amount of information to be learned, which helps us specify the criterion of a satisficing solution. Then the Blahut-Arimoto algorithm is employed to compute the target solution that reaches the minimum information rate under the distortion limit at each step. This method allows us to quickly identify a satisfactory solution in the least amount of time. To further enhance the time efficiency of the exploration, we propose a parallel version of the satisficing Thompson sampling strategy, which includes both synchronous and asynchronous versions. Furthermore, we have theoretically derived both the discounted and undiscounted cumulative regret bounds for our parallel methods.

The main contributions of our work can be summarized as:

(i) At the algorithmic level, by introducing the rate-distortion theory, we propose satisficing Thompson sampling-based parallel Bayesian optimization approaches, including synchronous and asynchronous versions. We consider time sensitivity and introduce the satisficing strategy that can accelerate the exploration process by allowing sub-optimality to Bayesian optimization.

(ii) At the theoretical level, we provide discounted and undiscounted regret bounds for the proposed approaches and analyze how our parallel approaches accelerate the exploration process compared to sequential methods in terms of regret bounds and loss function.

(iii) At the empirical level, the experimental results are accordant with theoretic analysis, and show that our parallel approaches outperform sequential methods and parallel BO using Thompson sampling in both synchronous and asynchronous settings.

The rest of this article is organized as follows. Section 2 describes the problem formulation for this work. In Section 3, we briefly revisit conventional Bayesian optimization approaches, including Gaussian progress regression and Thompson sampling. The STS-PBO method is put forward in Section 4. Section 5 provides discounted and undiscounted regret bounds for the proposed methods. Section 6 demonstrates the effectiveness of STS-PBO through the simulation experiment of fast charging design, followed by conclusions in Section 7.

## 2. Problem formulation

In Bayesian optimization problems, we would like to maximize or minimize a black-box function $f: \mathcal{X} \to \mathbb{R}$ defined on $\mathcal{X} \subset \mathbb{R}^d$. In this work, we take maximizing $f$ as an example and suppose that $\mathcal{X}$ is a finite set. At step $t$, we sample a data point $x_t$ and obtain a noisy observation:

$$y_t = f(\boldsymbol{x}_t) + \varepsilon_t, \tag{1}$$

where the noise $\varepsilon_t$ follows $\varepsilon_t \sim N(0, \eta^2)$.

Let $\boldsymbol{x}_* = \underset{x \in \mathcal{X}}{\operatorname{argmax}} f(\boldsymbol{x})$ denote the optimal solution. Recall that this work focuses on time-sensitivity optimization problems, in which each evaluation is time-intensive. Moreover, the parameter space we need to explore is extremely large, while our resources for such exploration are quite limited. In cases where the parameter space greatly exceeds the allowable number of evaluations, aiming for a satisficing solution often outperforms seeking for the optimal one. As such, our objective switches from the optimal solution $\boldsymbol{x}_*$ to a satisficing solution, denoted as $\widetilde{\boldsymbol{x}}$.

## 3. Bayesian optimization revisited

Most BO methods aim to identify an optimal point $\boldsymbol{x}_*$. To find $\boldsymbol{x}_*$, BO builds a probabilistic surrogate model to estimate the black-box function $f$ by continuously updating posterior knowledge and using an acquisition function to determine the next evaluation point. A common choice for the surrogate model is Gaussian process. Expected improvement (EI) [24], upper confidence bounds (UCB) [25] and Thompson sampling are often adopted for acquisition functions.

*3.1 Gaussian process regression*

Gaussian processes (GPs) are distribution over functions determined by a mean and covariance function [26]. The mean function is commonly initialized to zero in lack of prior knowledge, and the covariance between $\boldsymbol{x}_i$ and $\boldsymbol{x}_j$ is typically represented by a kernel function $\kappa(\boldsymbol{x}_i, \boldsymbol{x}_j)$ [26]. For an objective function $f(\boldsymbol{x})$, we assume that $f \sim \mathcal{GP}(m_f, \kappa_f)$ where $m_f: \mathcal{X} \to \mathbb{R}$ represents the mean function of $f$ and $\kappa_f: \mathcal{X} \times$ $\mathcal{X} \to \mathbb{R}$ represents the (covariance) kernel function. Given a set of observations $D$ consisting of inputs $x = \{x_i\}_{i=1}^N$ and corresponding outputs $y = \{y_i\}_{i=1}^N$, the posterior distribution of $f(x_*)$ can be written as

$$f(x_*)|D \sim N\left(\mu_f(x_*|D), \sigma_f^2(x_*|D)\right) \quad (2)$$

where

$$\mu_f(x_*|D) = k_*^T(K + \eta^2 I_N)^{-1} y, \quad (3)$$

$$\sigma_f^2(x_*|D) = k_{*,*} - k_*^T(K + \eta^2 I_N)^{-1} k_*. \quad (4)$$

Here $k_* = [\kappa_f(x_1, x_*), \kappa_f(x_2, x_*), \cdots, \kappa_f(x_N, x_*)]^T$, $[K]_{i,j} = \kappa_f(x_i, x_j)$, and $k_{*,*} = \kappa_f(x_*, x_*)$ [26]. For the kernel function we often choose the squared exponential (SE) kernel or the Matern kernel [18].

*3.2 Thompson sampling*

Thompson sampling, also known as probability matching, is a widely used Bayesian method for constructing acquisition function. The algorithm randomly selects the optimal point based on the posterior probability $p(x_*|D_t)$ at each step, where $D_t = \{(x_i, y_i)\}_{i=1}^{t-1}$ represents the history of evaluations, and has advantages over other related algorithms such as EI and UCB methods [27].

For GPs, considering $p(x_*|D_t) = \int p(x_*|g) p(g|D_t) \, dg$, at step $t$, TS first samples $g$ from the posterior distribution of $f$ conditioned on $D_t$ and then selects $x_t = \underset{x \in \mathcal{X}}{\text{argmax}} \, g(x)$ as the subsequent evaluation point. The result of this evaluation is incorporated into $D_t$, expressed as $D_{t+1} = D_t \cup \{(x_t, y_t)\}$ [18].

However, at every iteration, TS chooses unexplored points in pursuit of the optimal solution. This approach can quickly become inefficient as the size of the parameter space $\mathcal{X}$ grows [7]. Therefore, for time sensitive problems with large parameter spaces, it's crucial to investigate an optimization method that prioritizes short-term performance and targets a satisficing solution with minimal information acquisition.

## 4. Parallel Bayesian optimization with satisficing Thompson sampling

*4.1 Satisficing TS via rate-distortion technique*

Unlike most BO literature which aims to the identification of an optimal solution, we focus on how to learn a near-optimal, satisficing solution in the shortest possible time. It is worth noting the criticality of time sensitivity in real-world applications of BO, such as the design of fast charging for Li-ion batteries. Given that experimental evaluations in these problems are typically time-intensive, and the parameter space to be optimized considerably exceeds the number of experimental evaluations allowed to conduct, our objective is to minimize the number of evaluations needed to attain a nearly optimal satisficing solution. From an information theory perspective, this implies reducing the amount of information required to identify the target solution from the back-box function.

This brings us to the definition of the satisficing solution. The notion of a satisficing solution arises from the idea that learning a sub-optimal point often demands less information compared to the optimal one. Similar to the work of Arumugam and Van Roy [7], we introduce the rate-distortion theory to balance the sub-optimality and the bits of information required for learning. Let $d(\tilde{x}, g|D_t) = \mathbb{E}\left[\left(f(x_*) - f(\tilde{x})\right)^2 | f = g, D_t\right]$ measure the sub-optimality of $\tilde{x}$ under the estimate $g$ and the mutual information $\mathbb{I}(\tilde{x}; f)$ represent the amount of information for learning $\tilde{x}$ from $f$.

Then we define the rate-distortion function for BO as:

$$\mathcal{R}(D) = \inf_{\tilde{x} \sim \tilde{\mathcal{X}}} \mathbb{I}(\tilde{x}; f) \tag{5}$$

where $\tilde{\mathcal{X}} = \{\tilde{x}: \mathbb{E}[d(\tilde{x}, f)] \leq D, \tilde{x} \perp D_t | f, \forall t\}$. Here $\tilde{x} \perp D_t | f$ indicates that $\tilde{x}$ is independent of $D_t$ given the full information about $f$. An ideal satisficing point $\tilde{x}$ can minimize the information rate $\mathbb{I}(\tilde{x}; f)$ while satisfying the distortion limit $D$. Obtaining the posterior probability $p(\tilde{x}|D_t)$, we can sample $x_t \sim p(\tilde{x}|D_t)$ like TS, which is called satisficing TS.

Blahut-Arimoto algorithm [14, 15] is a classical method for solving the constrained rate-distortion optimization problem. First, a constraint objective is modified to minimize the Lagrangian loss function which is described as [7]:

$$\mathcal{L}_\beta(\tilde{x}|D_t) = \mathbb{I}_t(\tilde{x}; f) + \beta \mathbb{E}_t\left[\left(f(x_*) - f(\tilde{x})\right)^2\right] \tag{6}$$

where $\mathbb{I}_t(X;Y) = \mathbb{I}(X;Y|D_t = D_t)$ and $\mathbb{E}_t(X) = \mathbb{E}(X|D_t)$ [7]. The first term is the information rate which represents the bits of required information for learning $\tilde{x}$. The second term is a metric of distortion through the expected square error scaled by a Lagrange multiplier $\beta \in \mathbb{R}_{\geq 0}$ which represents the rate-distortion trade-off. Then apply the Blahut-Arimoto algorithm [7,14] to learn the target point $\tilde{x}$ that minimizes Eq. (6). The algorithm samples a set of posterior estimates of the unknown function, and is initialized with probabilities of the target point $\tilde{p}_0$ that depend on the estimation. An iterative sequence of $\tilde{p}_1$, $\tilde{p}_2$, $\cdots$ that converges on $\tilde{p}_*$ ($\tilde{p}_*(x|g) = \mathbb{P}(\tilde{x} = x|f = g)$) is generated as the exploration proceeds. Each iteration comprises two steps. The first step is to compute the marginal probabilities of $\tilde{x}$:

$$\tilde{q}_k(x) = \mathbb{E}_t[\tilde{p}_k(x|f)] \; \forall x \in \mathcal{X}, \tag{7}$$

and the second step is to update the probabilities of target point that depends on an

estimate $g$ of $f$:

$$\tilde{p}_{k+1}(x|g) = \frac{\tilde{q}_k(x)\exp\left(-\beta\mathbb{E}_t\left[(f(x_*)-f(x))^2|f=g\right]\right)}{\sum_{x'\in\mathcal{X}}\tilde{q}_k(x')\exp\left(-\beta\mathbb{E}_t\left[(f(x_*)-f(x'))^2|f=g\right]\right)}. \tag{8}$$

Generally, we choose the uniform distribution for the initial probabilities $\tilde{p}_0(x|g)$. It is worth mentioning that $\beta$ represents the expected prioritization of minimizing information rate versus minimizing distortion, which is the role of $D$ in Eq. (5) [7]. The target point thus calculated can reach the limit of the rate-distortion function, which is largely better than a manually selected satisficing point.

*4.2 Parallel satisficing TS for Bayesian optimization*

In parallel BO settings, suppose there are $M$ workers available to evaluate the objective function at different points concurrently. These parallel settings can be categorized into the settings of synchronous and asynchronous, as depicted in Fig.1.

In the synchronous setting, the algorithm determines a batch of $M$ points using the acquisition function based on posterior knowledge. These points are then directly assigned to the $M$ workers. All workers initiate their evaluations at the same time. Once every worker finishes, the algorithm updates the surrogate model and the acquisition function. Subsequently, it identifies and recommends the next batch of evaluation points in a similar manner, and repeats the aforementioned steps until a satisficing solution is obtained or the experiment budget is exhausted. While all evaluations contribute to the updated posterior knowledge of $f$ for the next batch's selection, there might be idle periods for some workers due to the variability in evaluation time.

In contrast, the asynchronous setting operates differently. Once a worker completes its evaluation, the algorithm instantly updates the surrogate model and identifies the next evaluation point for that worker. This approach maximizes worker efficiency, ensuring they remain active without idle periods. However, it's crucial to note that when updating the surrogate model, $M-1$ points are still under evaluation, which means not all evaluation results of the selected points can be used.

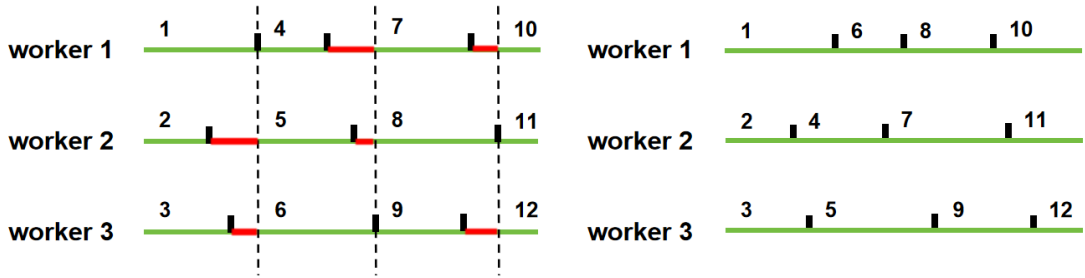

**Fig.1** A schematic comparison of the synchronous (left) and asynchronous (right) settings when $M=3$. The green horizontal line denotes an ongoing evaluation by a worker, while the black vertical line signifies its completion. In the synchronous setting, the red line on the left shows periods when the worker is idle. The numeral above the green line indicates the sequence of the evaluation.

In parallel BO settings, we have adapted the algorithm from Section 3.1 to propose the satisficing Thompson sampling-based parallel Bayesian optimization (STS-PBO) approaches, including both synchronous and asynchronous versions. In the asynchronous setting, when one worker finishes its evaluation and becomes available, the algorithm immediately updates the GP model and computes the next target point for the worker to start the next evaluation. The asynchronous version is presented in Algorithm 1. For the initial $M$ steps, since a worker is always idle, lines 3-5 are skipped. For the synchronous version, Algorithm 1 is adapted as follows: The algorithm pauses until all workers complete their tasks at line 3. Subsequently, these $M$ evaluations are used to update the evaluation results set and the posterior GP model in lines 4-5. After computing the posterior probabilities for the target point in lines 6-12, lines 13-14 are executed $M$ times to select the next batch of evaluation points. The algorithm then dispatches the $M$ workers to evaluate these points in line 15.

---

**Algorithm 1** The asynchronous STS-PBO approach

---

**Input**: Lagrange multiplier $\beta \in \mathbb{R}_{\geq 0}$, the number of Blahut-Arimoto iterations $K \in \mathbb{N}$, the number of posterior samples $Z \in \mathbb{N}$, the number of workers $M \in \mathbb{N}$

1: $D_1 = \{\}$, $\mathcal{GP}_1 = \mathcal{GP}(\mathbf{0}, \kappa)$
2: **for** $t = 1, 2, \cdots$ **do**
3:     Wait for a worker to finish
4:     $D_t = D_{t-1} \cup \{(\mathbf{x}', y')\}$ where $(\mathbf{x}', y')$ are the most recent observation of the worker
5:     Compute posterior $\mathcal{GP}_t$
6:     Sample $g_1, \cdots, g_Z \sim \mathcal{GP}_t$
7:     $d(\mathbf{x}, g_z | D_t) = \mathbb{E}\left[\left(f(\mathbf{x}_*) - f(\mathbf{x})\right)^2 | f = g_z, D_t\right], \forall \mathbf{x} \in \mathcal{X}, z \in Z$
8:     $\tilde{p}_0(\mathbf{x} | g_z) = \frac{1}{|\mathcal{X}|} \forall \mathbf{x} \in \mathcal{X}, z \in Z$
9:     **for** $k = 0, 1, \cdots, K-1$ **do**
10:         $\tilde{q}_k(\mathbf{x}) = \mathbb{E}_t[\tilde{p}_k(\mathbf{x}|f)] \, \forall \mathbf{x} \in \mathcal{X}$
11:         $\tilde{p}_{k+1}(\mathbf{x} | g_z) = \frac{\tilde{q}_k(\mathbf{x}) \exp(-\beta d(\mathbf{x}, g_z | D_t))}{\sum_{\mathbf{x}' \in \mathcal{X}} \tilde{q}_k(\mathbf{x}') \exp(-\beta d(\mathbf{x}', g_z | D_t))} \, \forall \mathbf{x} \in \mathcal{X}, \forall z \in Z$
12:     **end for**
13:     Sample $\hat{z} \sim \text{Uniform}(Z)$
14:     Sample $\mathbf{x}_t \sim \tilde{p}_K(\mathbf{x} | g_{\hat{z}})$
15:     Re-deploy the worker to evaluate the corresponding result of $\mathbf{x}_t$
16: **end for**

---

## 5. Theoretical analysis

To assess the performance of BO approaches, in line with many studies in BO literature, we use two metrics: simple regret (SR) and cumulative regret (CR), which are defined as

$$SR(\tau) = f(\mathbf{x}_*) - \underset{t=1,\cdots,\tau}{\mathrm{argmax}} f(\mathbf{x}_t) \tag{9}$$

$$CR(\tau) = \sum_{t=1}^{\tau}(f(x_*) - f(x_t)) \tag{10}$$

Additionally, to provide a thorough theoretical analysis highlighting the advantages of the proposed approach, we introduce both the undiscounted and the discounted Bayesian cumulative regret that are respectively defined as $\mathbb{E}[\sum_{t=\tau}^{\tau+T} f(x_*) - f(x_t) | D_\tau]$ and $\mathbb{E}[\sum_{t=\tau}^{\infty} \gamma^{t-\tau}(f(x_*) - f(x_t)) | D_\tau]$, where $\gamma \in [0,1)$ is a discount factor representing the preference for the performance of near-term versus long-term. This section adopts the two regret metrics for our theoretical analysis.

We now present both discounted and undiscounted Bayesian cumulative regret bounds for our proposed methods, referring to the work by Arumugam and Van Roy [7]. Owing to space limitations, detailed proofs of these results are provided in the appendix.

*5.1 Regret analysis of the synchronous version of STS-PBO algorithm*

Recall that the synchronous STS-PBO algorithm selects and evaluates a batch of points during each evaluation round. Specifically, the algorithm selects $x_{M(t-1)+1}, x_{M(t-1)+2}, \cdots, x_{Mt}$ based on the data $D_{t-syn} = (x_1, y_1, x_2, y_2, \cdots, x_{M(t-1)}, y_{M(t-1)})$, where $M$ denotes the number of workers. Therefore, we modify the loss function given by Eq. (6) as:

$$\mathcal{L}_{syn,\beta}(\tilde{x}|D_{t-syn}) = \mathbb{I}_{t-syn}(\tilde{x}; f) + \beta \mathbb{E}_{t-syn}\left[(f(x_*) - f(\tilde{x}))^2\right] \tag{11}$$

where $\mathbb{I}_{t-syn}(X;Y) = \mathbb{I}(X;Y|D_{t-syn} = D_{t-syn})$ and $\mathbb{E}_{t-syn}(X) = \mathbb{E}(X|D_{t-syn})$.

Let $(X_t, Y_t) = (x_{M(t-1)+m}, y_{M(t-1)+m})_{m=1}^{M}$, then we have the following lemma:

**Lemma 1.** For all $\beta > 0$, target points $\tilde{x}$ and $t = 1, 2, \cdots$, we have:

$$\mathbb{E}_{t-syn}[\mathcal{L}_{syn,\beta}(\tilde{x}|D_{(t+1)-syn})] = \mathcal{L}_{syn,\beta}(\tilde{x}|D_{t-syn}) - \mathbb{I}_{t-syn}(\tilde{x}; (X_t, Y_t)). \tag{12}$$

Let $\tilde{x}_{t-syn}$ be the target point conditioned on $D_{t-syn}$. Recall that by definition $\tilde{x}_{t-syn}$ minimizes $\mathcal{L}_{syn,\beta}(\tilde{x}_{t-syn}|D_{t-syn})$, and $x_{M(t-1)+1}, x_{M(t-1)+2}, \cdots, x_{Mt}$ are independently sampled from $\mathbb{P}(\tilde{x}_{t-syn} = \cdot | D_{t-syn})$. Based on these conditions, we obtain:**Lemma 2.** For all $\beta > 0$, target points $\tilde{x}$ and $t = 1, 2, \cdots$, we have:

$$\mathbb{E}_{t-syn}[\mathcal{L}_{syn,\beta}(\tilde{x}_{(t+1)-syn}|D_{(t+1)-syn})] \leq \mathcal{L}_{syn,\beta}(\tilde{x}_{t-syn}|D_{t-syn}) - \mathbb{I}_{t-syn}(\tilde{x}_{t-syn}; (X_t, Y_t)). \quad (13)$$

Then we can further derive the following corollary:

**Corollary 1.** For all $\beta > 0$, target points $\tilde{x}$ and $\tau = 1, 2, \cdots$, we have:

$$\mathbb{E}_{\tau-syn}\left[\sum_{t=\tau}^{\infty} \sum_{m=1}^{M} \mathbb{I}_{t-syn}(\tilde{x}_{t-syn}; (X_t, Y_t))\right] \leq M\mathcal{L}_{syn,\beta}(\tilde{x}|D_{\tau-syn}), \quad (14)$$

$$\mathbb{E}_{\tau-syn}\left[\sum_{t=\tau}^{T+\tau} \sum_{m=1}^{M} \mathbb{I}_{t-syn}(\tilde{x}_{t-syn}; (X_t, Y_t))\right] \leq M\mathcal{L}_{syn,\beta}(\tilde{x}|D_{\tau-syn}). \quad (15)$$

We modify the constant $\Gamma_{syn}$ to satisfy:

$$\Gamma_{syn} \geq \frac{\mathbb{E}_{t-syn}[f(\tilde{x}_t) - f(x_{M(t-1)+m})]^2}{\mathbb{I}_{t-syn}(\tilde{x}_t; (X_t, Y_t))}, \forall D_{t-syn}, m = 1, 2, \cdots, M. \quad (16)$$

Assume that the discount factor is for a single point rather than a batch of $M$ points. Then we obtain the following two regret bounds.

**Theorem 1.** (Discounted regret for the synchronous setting) If $\beta = \frac{1-\gamma^2}{(1-\gamma)^2 M\Gamma_{syn}}$, for all target points $\tilde{x}$ and $\tau = 1, 2, \cdots$, we have:

$$\mathbb{E}_{\tau-syn}\left[\sum_{t=\tau}^{\infty} \sum_{m=1}^{M} \gamma^{M(t-\tau)+m-1}\left(f(x_*) - f(x_{M(t-1)+m})\right)\right] \leq 2\sqrt{\frac{M\Gamma_{syn}\mathbb{I}_{\tau-syn}(\tilde{x}; f)}{1-\gamma^2}} + \frac{2\epsilon}{1-\gamma}, \quad (17)$$

where $\epsilon = \sqrt{\mathbb{E}_{\tau-syn}\left[(f(x_*) - f(\tilde{x}))^2\right]}$.

**Theorem 2.** (Undiscounted regret for the synchronous setting) If $\beta = \frac{T}{\Gamma_{syn}}$, for all target points $\tilde{x}$ and $\tau = 1, 2, \cdots$, we have:

$$\mathbb{E}_{\tau-syn}\left[\sum_{t=\tau}^{T+\tau} \sum_{m=1}^{M} f(x_*) - f(x_{M(t-1)+m})\right] \leq \quad (18)$$

$$2\sqrt{M\Gamma_{syn}MT\mathbb{I}_{\tau-syn}(\widetilde{\boldsymbol{x}};f)} + 2MT\epsilon,$$

where $\epsilon = \sqrt{\mathbb{E}_{\tau-syn}\left[(f(\boldsymbol{x}_*) - f(\widetilde{\boldsymbol{x}}))^2\right]}$.

The difference between $\beta$ in the synchronous method and the sequential method lies in the mutual information within $\Gamma_{syn}$. In the former, this information pertains to a batch of $M$ points, not just a single point as in the latter. This distinction introduces an additional factor of $M$ in the result of $\Gamma_{syn}$. Furthermore, the undiscounted regret for the synchronous method (i.e., Eq. (18)) considers $M(T+1)$ points, in contrast to the $T+1$ points considered in the undiscounted regret for the sequential setting. This introduces a factor of $M$ times $T$ in Eq. (18).

Compared with the regret bounds of the sequential approach, the synchronous algorithm benefits from a more extensive evaluation history after the same number of evaluation rounds ($D_{t-syn}$ versus $D_t$), which provides it with richer information about the target points. Moreover, regarding the loss function, Eq. (12) suggests that the loss function of synchronous method decreases in the mutual information of batch evaluation results, offering a faster decline than its sequential counterpart.

*5.2 Regret analysis of the asynchronous version of STS-PBO algorithm*

The asynchronous STS-PBO algorithm selects a new point as soon as a worker finishes its evaluation, even while there are still $M-1$ points awaiting evaluation. Specifically, the algorithm selects $\boldsymbol{x}_t$ based on the data $D_{t-asy} = (\boldsymbol{x}'_1, y'_1, \boldsymbol{x}'_2, y'_2, \cdots, \boldsymbol{x}'_{t-M}, y'_{t-M})$, where $(\boldsymbol{x}_t, y_t)$ denotes the $t$-th selected point and its corresponding evaluation result, and $(\boldsymbol{x}'_t, y'_t)$ denotes the $t$-th completed point and its corresponding evaluation result. Let the subscript $t - asy$ stand for $t -$

$M + 1$ in the evaluation results, and the loss function in Eq. (6) can be modified as

$$\mathcal{L}_{asy,\beta}(\tilde{x}|D_{t-asy}) = \mathbb{I}_{t-asy}(\tilde{x}; f) + \beta \mathbb{E}_{t-asy}\left[(f(x_*) - f(\tilde{x}))^2\right]. \tag{19}$$

where $\mathbb{I}_{t-asy}(X; Y) = \mathbb{I}(X; Y|D_{t-asy} = D_{t-asy})$ and $\mathbb{E}_{t-asy}(X) = \mathbb{E}(X|D_{t-asy})$. Then we have the following lemma:

**Lemma 3.** For all $\beta > 0$, target points $\tilde{x}$ and $t = M, M + 1, \cdots$, we have:

$$\mathbb{E}_{t-asy}[\mathcal{L}_{asy,\beta}(\tilde{x}|D_{(t+1)-asy})] = \mathcal{L}_{asy,\beta}(\tilde{x}|D_{t-asy}) - \mathbb{I}_{t-asy}\left(\tilde{x}; (x'_{t-asy}, y'_{t-asy})\right). \tag{20}$$

Recall that by definition $\tilde{x}_{t+1}$ minimizes $\mathcal{L}_{asy,\beta}(\tilde{x}_{t+1}|D_{(t+1)-asy})$, we can obtain:

**Lemma 4.** For all $\beta > 0$, target points $\tilde{x}$ and $t = M, M + 1, \cdots$, we have:

$$\mathbb{E}_{t-asy}[\mathcal{L}_{asy,\beta}(\tilde{x}_{t+1}|D_{(t+1)-asy})] \leq \mathcal{L}_{asy,\beta}(\tilde{x}_t|D_{t-asy}) - \mathbb{I}_{t-asy}\left(\tilde{x}_t; (x'_{t-asy}, y'_{t-asy})\right). \tag{21}$$

Therefore, we can further derive the following corollary:

**Corollary 2.** For all $\beta > 0$, target points $\tilde{x}$ and $\tau = M, M + 1, \cdots$, we have:

$$\mathbb{E}_{\tau-asy}\left[\sum_{t=\tau}^{\infty} \mathbb{I}_{t-asy}\left(\tilde{x}_t; (x'_{t-asy}, y'_{t-asy})\right)\right] \leq \mathcal{L}_{asy,\beta}(\tilde{x}|D_{\tau-asy}), \tag{22}$$

$$\mathbb{E}_{\tau-asy}\left[\sum_{t=\tau}^{T+\tau} \mathbb{I}_{t-asy}\left(\tilde{x}_t; (x'_{t-asy}, y'_{t-asy})\right)\right] \leq \mathcal{L}_{asy,\beta}(x|D_{\tau-asy}). \tag{23}$$

We modify the constant $\Gamma_{asy}$ to satisfy

$$\Gamma_{asy} \geq \frac{\mathbb{E}_{t-asy}[f(\tilde{x}_t) - f(x_t)]^2}{\mathbb{I}_{t-asy}(\tilde{x}_t; (x'_{t-asy}, y'_{t-asy}))}, \forall D_{t-asy}, \tag{24}$$

where $x_t$ is independently sampled from the marginal distribution of $\tilde{x}_t$. Then we obtain the following two regret bounds.

**Theorem 3.** (Discounted regret for the asynchronous setting) If $\beta = \frac{1-\gamma^2}{(1-\gamma)^2 \Gamma_{asy}}$, for all target points $\tilde{x}$ and $\tau = M, M + 1, \cdots$, we have:

$$\mathbb{E}_{\tau-asy}\left[\sum_{t=\tau}^{\infty} \gamma^{t-\tau}(f(x_*) - f(x_t))\right] \leq 2\sqrt{\frac{\Gamma_{asy} \mathbb{I}_{\tau-asy}(x; f)}{1-\gamma^2}} + \frac{2\epsilon}{1-\gamma}, \tag{25}$$

where $\epsilon = \sqrt{\mathbb{E}_{\tau-asy}\left[(f(\boldsymbol{x}_*) - f(\widetilde{\boldsymbol{x}}))^2\right]}$.

**Theorem 4.** (Undiscounted regret for the asynchronous setting) If $\beta = \frac{T}{\Gamma_{asy}}$, for all target points $\widetilde{\boldsymbol{x}}$ and $\tau = M, M+1, \cdots$, we have:

$$\mathbb{E}_{\tau-asy}[\sum_{t=\tau}^{T+\tau} f(\boldsymbol{x}_*) - f(\boldsymbol{x}_t)] \leq 2\sqrt{\Gamma_{asy}T\mathbb{I}_{\tau-asy}(\boldsymbol{x}; f)} + 2T\epsilon, \tag{26}$$

where $\epsilon = \sqrt{\mathbb{E}_{\tau-asy}\left[(f(\boldsymbol{x}_*) - f(\widetilde{\boldsymbol{x}}))^2\right]}$.

Notably, in Eq. (24), there is a gap of $M-1$ points in the mutual information of the denominator. Compared with the regret bounds of its sequential counterpart, the asynchronous algorithm, despite the information gap of $M-1$ points, can explore more points (a larger $\tau$) within the same timeframe. This provides the asynchronous method with more information about the target points. Additionally, a comparison between Eq. (20) and its sequential version shows that the asynchronous method accelerates the reduction of the loss function by increasing mutual information drops in the same timeframe, which aligns well with our intuitive expectations. However, when comparing the performance of asynchronous and synchronous algorithms, it's crucial to weigh the effects of both the $M-1$ point information gap and the number of evaluations.Given that asynchronous algorithm typically undertakes more evaluations than its synchronous counterpart (often by more than $M-1$), the performance of the asynchronous method tends to be superior, especially when evaluation times exhibit significant variability.

## 6. Simulation study

In this section, a simulation experiment of fast-charging design problem for Lithium-ion batteries [2, 29, 30] is used to evaluate the performance of our proposed algorithms. The objective of fast-charging design is to optimize a charging strategy that minimizes battery degradation (i.e., maximizes battery lifespan) with a fixed charging duration. Here, the lifespan denotes the number of charge-discharge cycles until the battery capacity degrades to 80% of its initial capacity.

The charging strategies in our experiments follow the commonly used constant current-constant voltage (CC-CV) charging protocol [32]. We partition the fix charging duration into $n$ time intervals. Within each interval, a constant charging current $I_j$ is applied during the $j$-th interval. Let's represent a fast-charging strategy as $x = (I_1, I_2, \cdots, I_n)$. Thus, for a given $n$, the fast-charging optimization problem can be formulated as:

$$x_* = \underset{x=(I_1,I_2,\cdots,I_n)\in \mathcal{X}}{\mathrm{argmax}} f(x), \qquad (27)$$

where $\mathcal{X} \in \mathbb{R}^n$ represents a set of charging protocol that meets the fixed charging time and $f: \mathcal{X} \to \mathbb{R}$ represents a mapping from charging protocol to battery cycle life (here $f$ is a black-box function). In this work, the charging current is divided into three steps (i.e., $n = 3$), denoted as $I_1$, $I_2$, and $I_3$, where $I_1$ and $I_2$ are free variables and $I_3$ is determined by the preceding currents. We denote the change in the state of charge (SOC) before and after each charging step as $\Delta Q_i, i = 1,2,3$. Similarly, the duration of each step is represented as $\Delta t_i, i = 1,2,3$, which can be computed as

$$t_i = \frac{\Delta Q_i}{I_i}, i = 1,2, \quad t_3 = t_f - t_1 - t_2, \quad I_3 = \frac{\Delta Q_3}{t_3} \qquad (28)$$

where, $t_f$ is the total charging time required to charge the SOC from 0% to 80%, and $t_f$=800 s is used in this work. The currents $I_1$, $I_2$, and $I_3$ correspond to the SOC increase from 0% to 20%, 20% to 40%, and 40% to 80%, respectively. The ranges for both $I_1$ and $I_2$ span from 2.2 C to 6.0 C, incremented in steps of 0.2 C.

We split the ranges of $I_1$ and $I_2$ into fine grids and simulate the battery lifetime for each grid. The battery lifetime across this two-dimensional spectrum of current values is shown in Fig. 2. More information on the fast-charging design problem and the associated dataset can be found in [2,30].

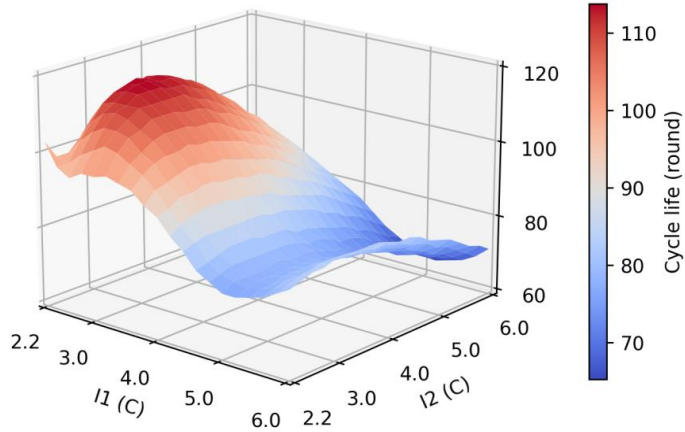

**Fig.2.** Battery lifetime across the two-dimensional spectrum of current values for the simulated charging strategies

In all of our experiments, we set the noise-to-signal ratio to be 5%, namely, $\varepsilon(x) \sim N(0, \sigma(x)^2)$, and $\sigma(x) = 0.05 f(x)$). For the STS-PBO algorithm, the number of posterior samples $Z$ is set to 64 and the maximum number of iterations $K$ is set to 100, stopping early if the difference between two consecutive iterations falls below a threshold [7]. The hyper-parameter $\beta$ is selected from $\{0.01, 0.05, 0.1, 1\}$. For each algorithm setting, we use 20 random seeds to improve the reliability of the results.

Considering that the time for selecting a strategy is far shorter than the time for evaluating the strategy, we ignore the selection time and take the total evaluation time as the criterion. The evaluation time of a strategy is proportional to the cycle life under the strategy. Therefore, we use rounds as the unit of evaluation time. In all of our experiments, we set the maximum total evaluation time to 10000 rounds (approximately 100 evaluations under sequential methods).

To evaluate the performance of our proposed algorithm, we select two metrics: (1) mean regret and (2) minimum regret. The mean regret is the mean of the cumulative regret in Eq. (10) (i.e., $CR(\tau)/\tau$) and the minimum regret is the simple regret ($SR(\tau)$) in Eq. (9). Here we use mean regret instead of cumulative regret because the number of evaluations is different for sequential and asynchronous methods within the same time and we would like to compare the performance of these methods in the same total time.

*6.1 Results and discussion*

We validate our proposed STS-PBO algorithm compared with (1) the TS-BO algorithm, (2) the STS-BO algorithm, and (3) the TS-PBO algorithm. The mean regret results and minimum regret results are depicted in Fig. 3 and 4. For the mean regret results, we can see from Fig.3 that the parallel algorithm significantly outperforms the sequential algorithm both for TS-BO and STS-BO as we expected. Moreover, in parallel algorithms, the asynchronous version performs as well as or slightly better than the synchronous version. From another perspective, compared with TS-(P)BO, STS-(P)BO can achieve a smaller mean regret over the same total exploration time, especially when $\beta$ is small. When $\beta$ is small, the mean regret that can be achieved within the same total evaluation time is also small. As $\beta \to \infty$, however, STS-BO using the Blahut-Arimoto algorithm recovers the performance of TS-BO gradually. The reason is that, as shown in (6), the objective of the STS-BO algorithm is concentrated on minimizing distortion as $\beta$ tends to $\infty$, which recovers the target of Thompson sampling. We also find that this conclusion exists not only in the sequential version, but also in the synchronous and asynchronous version. In summary, the asynchronous STS-PBO algorithm performs best in terms of mean regret.

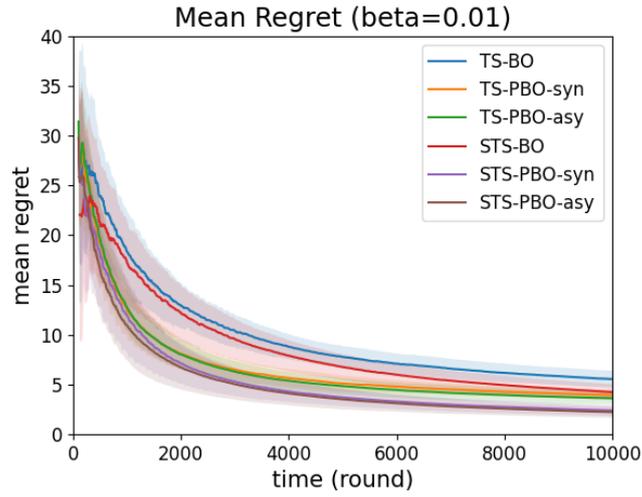

(a)

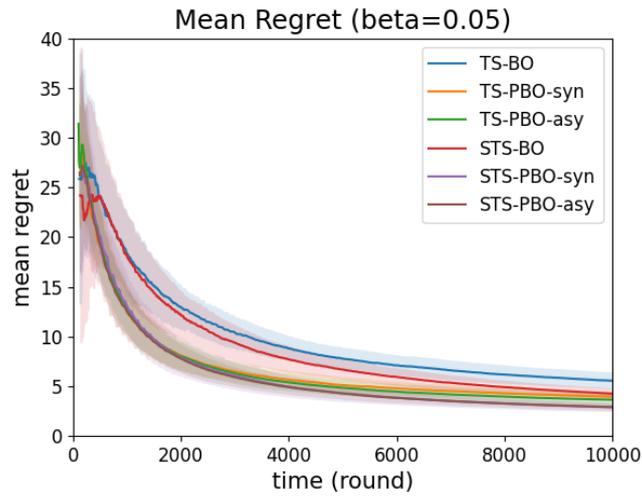

(b)

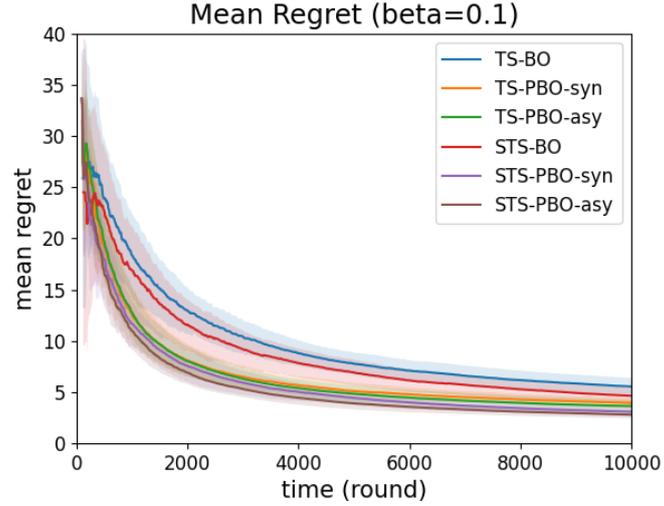

(c)

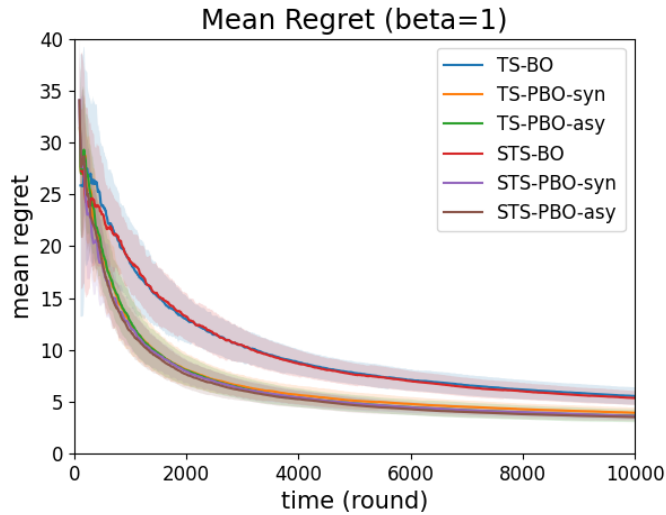

(d)

**Fig3.** The mean and standard deviation of the mean regret of cycle life by using the methods of (1) TS-BO, (2) TS-PBO-syn (synchronous), (3) TS-PBO-asy (asynchronous), (4) STS-BO, (5) STS-PBO-syn, and (6) STS-PBO-asy. Each method is repeated 20 times. The hyper-parameter $\beta$ is set at four different values: (a) 0.01, (b) 0.05, (c) 0.1, and (d) 1.

For the minimum regret results, to compare the performance of different methods more clearly, we intercept the results of the first 2000 rounds (approximately 20 evaluations under sequential methods). We can see from Fig. 4 that both for TS-(P)BO and STS-(P)BO the parallel method performs better than the sequential method. In parallel methods, the asynchronous version also performs as well as or slightly better than the synchronous version. However, when comparing TS-(P)BO and STS-(P)BO, we need to consider the value of $\beta$. When $\beta$ is small ($\beta = 0.01$), the performance of STS-(P)BO algorithm is slightly inferior to TS-(P)BO algorithm. When the value of $\beta$ is appropriate ($\beta = 0.05 / 0.1$), the performance of STS-(P)BO is as good as or slightly better than that of TS-(P)BO. When the value of $\beta$ increases ($\beta = 1$), STS-(P)BO recovers the performance of TS-(P)BO. That is because when $\beta$ is small, greater regret is allowed and STS-(P)BO behaves like the uniform random policy. As $\beta$ grows, permitted regret decreases and STS-(P)BO behaves like TS-(P)BO. In summary, when the value of $\beta$ is appropriate, the asynchronous STS-PBO algorithm performs best in terms of minimum regret in our experiment.

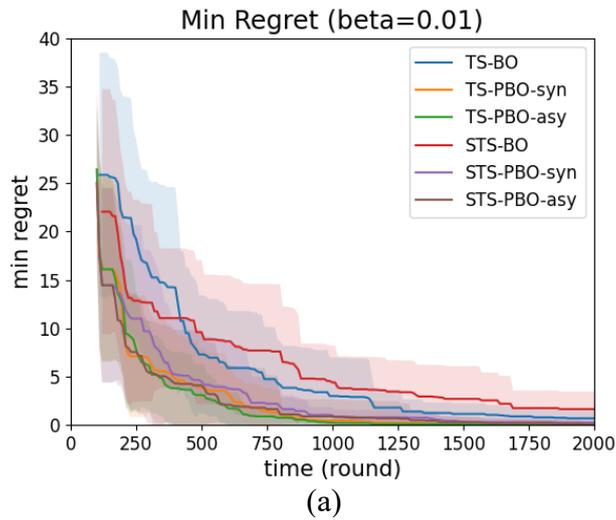

(a)

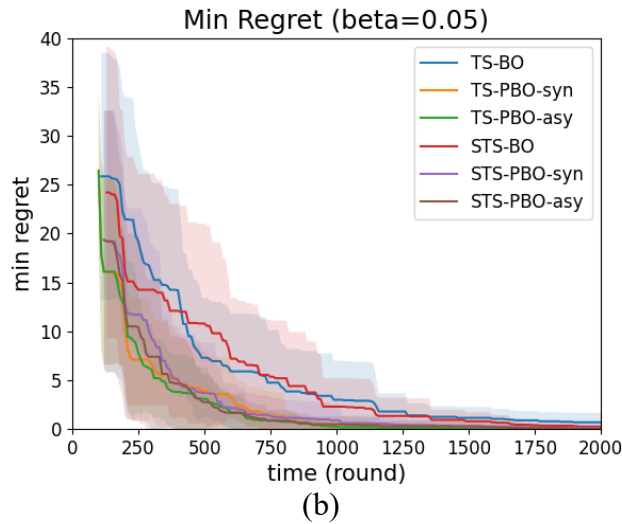

(b)

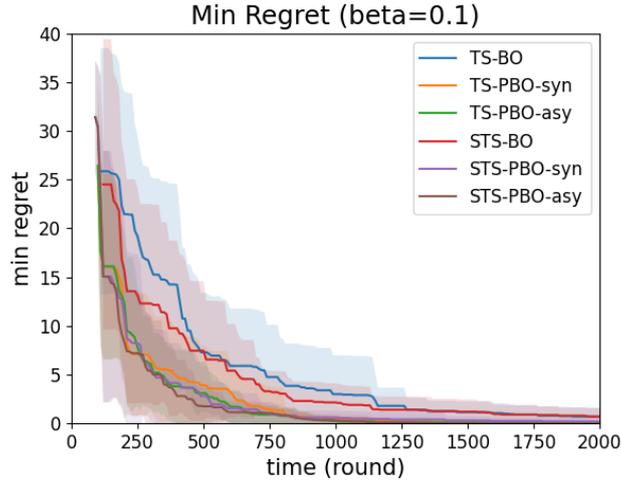

(c)

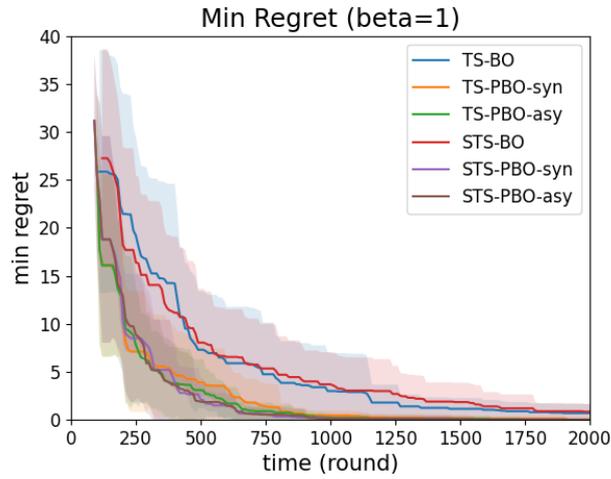

(d)

**Fig4.** The mean and standard deviation of the minimum regret of cycle life by using the methods of (1) TS-BO, (2) TS-PBO-syn, (3) TS-PBO-asy, (4) STS-BO, (5) STS-PBO-syn, and (6) STS-PBO-asy. Each method is repeated 20 times. The hyper-parameter $\beta$ is set at four different values: (a) 0.01, (b) 0.05, (c) 0.1, and (d) 1. For a clearer comparison of the performances across different methods, we focus on the results from the initial 2000 rounds.

It is worth mentioning that neither mean regret nor minimum regret alone can fully reflect the exploration and exploiting ability of the algorithm. For mean regret, even if the mean regret reaches a very low level, there is no guarantee that the algorithm has identified the satisficing strategy. For minimum regret, even if the optimal strategy in theory has been found, there is no guarantee that the algorithm will always identify it as the optimal action due to the presence of noise. We need to take the two metrics into consideration comprehensively. In our experiments, we adopt both of the two metrics and draw our conclusions that our proposed STS-PBO methods can improve the performance of the exploration and exploitation in the fast charging problems when the value of $\beta$ is chosen appropriately, especially the asynchronous STS-PBO algorithm.

## 7. Conclusions

In this article, to tackle time-sensitivity black-box optimization problems, we propose satisficing Thompson sampling-based parallel Bayesian optimization approaches, including synchronous and asynchronous versions. Instead of seeking an optimal solution, our objective shifts towards identifying a satisficing solution that allows sub-optimality to reduce the bits of information that an agent needs to learn. The Blahut-Arimoto algorithm from rate-distortion theory is adopted to calculate this satisficing solution which minimizes the amount of information to be learned while satisfying the sub-optimality constraint. Moreover, we incorporate parallel strategies that enhance exploration and exploitation efficiency by increasing the number of evaluations within the same total evaluation time. We provide both discounted and undiscounted regret bounds for our proposed methods, and delve into the advantages of parallel methods compared with the sequential counterparts in terms of regret bounds and loss function. The efficacy of the proposed STS-PBO methods is demonstrated on a simulation experiment of fast-charging design problem for Lithium-ion batteries The simulation results show that STS-PBO methods outperform their counterparts in both mean and minimum regrets when the value of $\beta$ is appropriate. As a direction for future work, the method for the self-adaptive update of the hyper-parameter $\beta$ can be employed to prevent inappropriate hyper-parameter selection that may hamper the performance of our proposed approaches [7].

**Appendix. Proofs of regret bounds**

*A.1. Proofs of regret bounds for the synchronous version of STS-PBO*

**Lemma 1.** For all $\beta > 0$, target points $\tilde{x}$ and $t = 1, 2, \cdots$, we have:

$$\mathbb{E}_{t-syn}[\mathcal{L}_{syn,\beta}(\tilde{x}|D_{(t+1)-syn})] = \mathcal{L}_{syn,\beta}(\tilde{x}|D_{t-syn}) - \mathbb{I}_{t-syn}(\tilde{x}; (X_t, Y_t)).$$

*Proof*: Recall that $D_{(t+1)-syn} = (D_{t-syn}, X_t, Y_t)$. As mentioned in section 3.1, we have $\tilde{x} \perp D_t | f, \forall t$. Therefore, we can obtain:

$$\mathbb{I}_{t-syn}((X_t, Y_t); \tilde{x}|f) = 0.$$

where $\mathbb{I}_t(X; Y|Z) = \mathbb{I}(X; Y|D_t = D_t, Z)$ [7]. Due to the chain rule of mutual information:

$$\mathbb{I}(X; Z_1, Z_2, \cdots, Z_n) = \sum_{i=1}^{n} \mathbb{I}(X; Z_i | Z_1, \cdots, Z_{i-1}),$$

we can obtain:

$$\mathbb{I}_{t-syn}(\tilde{x}; f) = \mathbb{I}_{t-syn}(\tilde{x}; f) + \mathbb{I}_{t-syn}(\tilde{x}; (X_t, Y_t)|f)$$

$$= \mathbb{I}_{t-syn}(\tilde{x}; f, (X_t, Y_t))$$

$$= \mathbb{I}_{t-syn}(\tilde{x}; f|(X_t, Y_t)) + \mathbb{I}_{t-syn}(\tilde{x}; (X_t, Y_t))$$

It follows that:

$$\mathbb{E}_{t-syn}[\mathcal{L}_{syn,\beta}(\tilde{x}|D_{(t+1)-syn})]$$

$$= \mathbb{E}[\mathcal{L}_{syn,\beta}(\tilde{x}|D_{(t+1)-syn})|D_{t-syn}]$$

$$= \mathbb{E}\left[\mathbb{I}_{t-syn}(\tilde{x}; f|(X_t, Y_t)) + \beta \mathbb{E}\left[(f(x_*) - f(\tilde{x}))^2 | D_{(t+1)-syn}\right] | D_{t-syn}\right]$$

$$= \mathbb{E}_{t-syn}[\mathbb{I}_{t-syn}(\tilde{x}; f|(X_t, Y_t))] + \beta \mathbb{E}_{t-syn}\left[(f(x_*) - f(\tilde{x}))^2\right]$$

$$= \mathbb{E}_{t-syn}[\mathbb{I}_{t-syn}(\tilde{x}; f) - \mathbb{I}_{t-syn}(\tilde{x}; (X_t, Y_t))] + \beta \mathbb{E}_{t-syn}\left[(f(x_*) - f(\tilde{x}))^2\right]$$

$$= \mathbb{I}_{t-syn}(\widetilde{\pmb{x}};\ f) + \beta\mathbb{E}_{t-syn}\left[(f(\pmb{x}_*) - f(\widetilde{\pmb{x}}))^2\right] - \mathbb{I}_{t-syn}(\widetilde{\pmb{x}};\ (\pmb{X}_t, Y_t))$$

$$= \mathcal{L}_{syn,\beta}(\widetilde{\pmb{x}}|D_{t-syn}) - \mathbb{I}_{t-syn}(\widetilde{\pmb{x}};\ (\pmb{X}_t, Y_t))$$

**Lemma 2.** For all $\beta > 0$, target points $\widetilde{\pmb{x}}$ and $t = 1, 2, \cdots$, we have:

$$\mathbb{E}_{t-syn}[\mathcal{L}_{syn,\beta}(\widetilde{\pmb{x}}_{(t+1)-syn}|D_{(t+1)-syn})] \leq \mathcal{L}_{syn,\beta}(\widetilde{\pmb{x}}_{t-syn}|D_{t-syn}) - \mathbb{I}_{t-syn}\left(\widetilde{\pmb{x}}_{t-syn};\ (\pmb{X}_t, Y_t)\right).$$

*Proof*: Recall that by definition $\widetilde{\pmb{x}}_{(t+1)-syn}$ minimizes $\mathcal{L}_{syn,\beta}(\widetilde{\pmb{x}}_{(t+1)-syn}|D_{(t+1)-syn})$, we then have:

$$\mathbb{E}\left[\mathcal{L}_{syn,\beta}(\widetilde{\pmb{x}}_{(t+1)-syn}|D_{(t+1)-syn})\middle|D_{t-syn}\right]$$

$$\leq \mathbb{E}\left[\mathcal{L}_{syn,\beta}(\widetilde{\pmb{x}}_{t-syn}|D_{(t+1)-syn})\middle|D_{t-syn}\right]$$

$$= \mathcal{L}_{syn,\beta}(\widetilde{\pmb{x}}_{t-syn}|D_{t-syn}) - \mathbb{I}_{t-syn}\left(\widetilde{\pmb{x}}_{t-syn};\ (\pmb{X}_t, Y_t)\right)$$

**Corollary 1.** For all $\beta > 0$, target points $\widetilde{\pmb{x}}$ and $\tau = 1, 2, \cdots$, we have:

$$\mathbb{E}_{\tau-syn}\left[\sum_{t=\tau}^{\infty}\sum_{m=1}^{M} \mathbb{I}_{t-syn}\left(\widetilde{\pmb{x}}_{t-syn};\ (\pmb{X}_t, Y_t)\right)\right] \leq M\mathcal{L}_{syn,\beta}(\widetilde{\pmb{x}}|D_{\tau-syn}),$$

$$\mathbb{E}_{\tau-syn}\left[\sum_{t=\tau}^{T+\tau}\sum_{m=1}^{M} \mathbb{I}_{t-syn}\left(\widetilde{\pmb{x}}_{t-syn};\ (\pmb{X}_t, Y_t)\right)\right] \leq M\mathcal{L}_{syn,\beta}(\widetilde{\pmb{x}}|D_{\tau-syn}).$$

*Proof*:

$$\mathbb{E}_{\tau-syn}\left[\sum_{t=\tau}^{\infty}\sum_{m=1}^{M} \mathbb{I}_{t-syn}\left(\widetilde{\pmb{x}}_{t-syn};\ (\pmb{X}_t, Y_t)\right)\right]$$

$$\leq \mathbb{E}_{\tau-syn}\left[\sum_{t=\tau}^{\infty}\sum_{m=1}^{M} \mathcal{L}_{syn,\beta}(\widetilde{\pmb{x}}_{t-syn}|D_{t-syn}) - \mathbb{E}_{t-syn}[\mathcal{L}_{syn,\beta}(\widetilde{\pmb{x}}_{(t+1)-syn}|D_{(t+1)-syn})]\right]$$

$$= \sum_{t=\tau}^{\infty}\sum_{m=1}^{M} \mathbb{E}_{\tau-syn}[\mathcal{L}_{syn,\beta}(\widetilde{\pmb{x}}_{t-syn}|D_{t-syn})]$$

$$\quad - \mathbb{E}_{\tau-syn}\left[\mathbb{E}_{t-syn}[\mathcal{L}_{syn,\beta}(\widetilde{\pmb{x}}_{(t+1)-syn}|D_{(t+1)-syn})]\right]$$

$$= \sum_{m=1}^{M} \mathbb{E}_{\tau-syn}[\mathcal{L}_{syn,\beta}(\widetilde{\boldsymbol{x}}_{\tau-syn}|D_{\tau-syn})] + \sum_{t=\tau+1}^{\infty} \sum_{m=1}^{M} \mathbb{E}_{\tau-syn}[\mathcal{L}_{syn,\beta}(\widetilde{\boldsymbol{x}}_{t-syn}|D_{t-syn})]$$

$$- \sum_{t=\tau}^{\infty} \sum_{m=1}^{M} \mathbb{E}_{\tau-syn}[\mathcal{L}_{syn,\beta}(\widetilde{\boldsymbol{x}}_{(t+1)-syn}|D_{(t+1)-syn})]$$

$$= \sum_{m=1}^{M} \mathcal{L}_{syn,\beta}(\widetilde{\boldsymbol{x}}_{\tau-syn}|D_{\tau-syn}) \leq M\mathcal{L}_{syn,\beta}(\widetilde{\boldsymbol{x}}|D_{\tau-syn})$$

where the proofs follow the fact that $\widetilde{\boldsymbol{x}}_{\tau-syn}$ minimizes $\mathcal{L}_{syn,\beta}(\widetilde{\boldsymbol{x}}_{\tau-syn}|D_{\tau-syn})$. Then we have:

$$\mathbb{E}_{\tau-syn}\left[\sum_{t=\tau}^{T+\tau} \sum_{m=1}^{M} \mathbb{I}_{t-syn}(\widetilde{\boldsymbol{x}}_{t-syn}; (\boldsymbol{X}_t, Y_t))\right]$$

$$\leq \mathbb{E}_{\tau-syn}\left[\sum_{t=\tau}^{\infty} \sum_{m=1}^{M} \mathbb{I}_{t-syn}(\widetilde{\boldsymbol{x}}_{t-syn}; (\boldsymbol{X}_t, Y_t))\right]$$

$$\leq M\mathcal{L}_{syn,\beta}(\widetilde{\boldsymbol{x}}|D_{\tau-syn})$$

**Theorem 1.** (Discounted regret for the synchronous setting) If $\beta = \frac{1-\gamma^2}{(1-\gamma)^2 M \Gamma_{syn}}$, for all target points $\widetilde{\boldsymbol{x}}$ and $\tau = 1, 2, \cdots$, we have:

$$\mathbb{E}_{\tau-syn}\left[\sum_{t=\tau}^{\infty} \sum_{m=1}^{M} \gamma^{M(t-\tau)+m-1}\left(f(\boldsymbol{x}_*) - f(\boldsymbol{x}_{M(t-1)+m})\right)\right] \leq 2\sqrt{\frac{M\Gamma_{syn}\mathbb{I}_{\tau-syn}(\widetilde{\boldsymbol{x}};f)}{1-\gamma^2}} + \frac{2\epsilon}{1-\gamma},$$

where $\epsilon = \sqrt{\mathbb{E}_{\tau-syn}\left[(f(\boldsymbol{x}_*) - f(\widetilde{\boldsymbol{x}}))^2\right]}$.

*Proof:* First, we have:

$$\mathbb{E}_{\tau-syn}\left[\sum_{t=\tau}^{\infty} \sum_{m=1}^{M} \gamma^{M(t-\tau)+m-1}\left(f(\boldsymbol{x}_*) - f(\boldsymbol{x}_{M(t-1)+m})\right)\right]$$

$$= \mathbb{E}_{\tau-syn}\left[\sum_{t=\tau}^{\infty} \sum_{m=1}^{M} \gamma^{M(t-\tau)+m-1}\left(f(\boldsymbol{x}_*) - f(\widetilde{\boldsymbol{x}}_{t-syn}) + f(\widetilde{\boldsymbol{x}}_{t-syn}) - f(\boldsymbol{x}_{M(t-1)+m})\right)\right]$$

$$= \mathbb{E}_{\tau-syn}\left[\sum_{t=\tau}^{\infty}\sum_{m=1}^{M}\gamma^{M(t-\tau)+m-1}\left(f(x_*)-f(\tilde{x}_{t-syn})\right)\right]$$

$$+ \mathbb{E}_{\tau-syn}\left[\sum_{t=\tau}^{\infty}\sum_{m=1}^{M}\gamma^{M(t-\tau)+m-1}\left(f(\tilde{x}_{t-syn})-f(x_{M(t-1)+m})\right)\right]$$

Given the definition of $\mathcal{L}_{syn,\beta}(\tilde{x}_{t-syn}|D_{t-syn})$ and the value of $\beta$, we have:

$$\mathcal{L}_{syn,\beta}(\tilde{x}_{t-syn}|D_{t-syn}) = \mathbb{I}_{t-syn}(\tilde{x}_{t-syn}; f) + \beta \mathbb{E}_{t-syn}\left[\left(f(x_*)-f(\tilde{x}_{t-syn})\right)^2\right]$$

$$\geq \frac{1-\gamma^2}{(1-\gamma)^2 M\Gamma_{syn}}\mathbb{E}_{t-syn}\left[\left(f(x_*)-f(\tilde{x}_{t-syn})\right)^2\right]$$

Using Jensen's inequality and the fact that $\tilde{x}_{t-syn}$ minimizes $\mathcal{L}_{syn,\beta}(\tilde{x}_{t-syn}|D_{t-syn})$, we have:

$$\mathbb{E}_{t-syn}[f(x_*)-f(\tilde{x}_{t-syn})] \leq \sqrt{\mathbb{E}_{t-syn}\left[\left(f(x_*)-f(\tilde{x}_{t-syn})\right)^2\right]}$$

$$\leq \sqrt{\frac{(1-\gamma)^2 M\Gamma_{syn}}{1-\gamma^2}\mathcal{L}_{syn,\beta}(\tilde{x}_{t-syn}|D_{t-syn})}$$

$$\leq (1-\gamma)\sqrt{\frac{M\Gamma_{syn}\mathcal{L}_{syn,\beta}(\tilde{x}_{t-syn}|D_{t-syn})}{1-\gamma^2}}$$

Lemma 1 implies that for all $t \geq \tau$:

$$\mathbb{E}_{\tau-syn}[\mathcal{L}_{syn,\beta}(\tilde{x}|D_{t-syn})] \leq \mathcal{L}_{syn,\beta}(\tilde{x}|D_{\tau-syn}).$$

Combined with Jensen's inequality, we have:

$$\mathbb{E}_{\tau-syn}[f(x_*)-f(\tilde{x}_{t-syn})]$$

$$\leq (1-\gamma)\mathbb{E}_{\tau-syn}\left[\sqrt{\frac{M\Gamma_{syn}\mathcal{L}_{syn,\beta}(\tilde{x}|D_{t-syn})}{1-\gamma^2}}\right]$$

$$\leq (1-\gamma)\sqrt{\frac{M\Gamma_{syn}\mathbb{E}_{\tau-syn}[\mathcal{L}_{syn,\beta}(\tilde{x}|D_{t-syn})]}{1-\gamma^2}}$$

$$\leq (1-\gamma)\sqrt{\frac{M\Gamma_{syn}\mathcal{L}_{syn,\beta}(\tilde{x}|D_{\tau-syn})}{1-\gamma^2}}$$

It follows that:

$$\mathbb{E}_{\tau-syn}\left[\sum_{t=\tau}^{\infty}\sum_{m=1}^{M}\gamma^{M(t-\tau)+m-1}\left(f(\boldsymbol{x}_*)-f(\widetilde{\boldsymbol{x}}_{t-syn})\right)\right]$$

$$\leq \sqrt{\frac{M\Gamma_{syn}\mathcal{L}_{syn,\beta}(\widetilde{\boldsymbol{x}}|D_{\tau-syn})}{1-\gamma^2}}$$

$$\leq \sqrt{\frac{M\Gamma_{syn}\left(\mathbb{I}_{\tau-syn}(\widetilde{\boldsymbol{x}};f)+\beta\epsilon^2\right)}{1-\gamma^2}}$$

$$\leq \sqrt{\frac{M\Gamma_{syn}\mathbb{I}_{\tau-syn}(\widetilde{\boldsymbol{x}};f)}{1-\gamma^2}}+\frac{\epsilon}{1-\gamma}$$

In addition, from the definition of $\Gamma_{syn}$, Corollary 1, and the Cauchy-Schwartz inequality, we have:

$$\mathbb{E}_{\tau-syn}\left[\sum_{t=\tau}^{\infty}\sum_{m=1}^{M}\gamma^{M(t-\tau)+m-1}\left(f(\widetilde{\boldsymbol{x}}_{t-syn})-f(\boldsymbol{x}_{M(t-1)+m})\right)\right]$$

$$\leq \mathbb{E}_{\tau-syn}\left[\sum_{t=\tau}^{\infty}\sum_{m=1}^{M}\gamma^{M(t-\tau)+m-1}\sqrt{\Gamma_{syn}\mathbb{I}_{t-syn}\left(\widetilde{\boldsymbol{x}}_{t-syn};(\boldsymbol{X}_t,Y_t)\right)}\right]$$

$$\leq \sqrt{\sum_{t=\tau}^{\infty}\sum_{m=1}^{M}\gamma^{2M(t-\tau)+2m-2}\Gamma_{syn}\sum_{t=\tau}^{\infty}\sum_{m=1}^{M}\mathbb{E}_{\tau-syn}\left[\mathbb{I}_{t-syn}\left(\widetilde{\boldsymbol{x}}_{t-syn};(\boldsymbol{X}_t,Y_t)\right)\right]}$$

$$\leq \sqrt{\Gamma_{syn}M\mathcal{L}_{syn,\beta}(\widetilde{\boldsymbol{x}}|D_{\tau-syn})\sum_{t=0}^{\infty}\gamma^{2t}}$$

$$= \sqrt{\frac{\Gamma_{syn}M\mathcal{L}_{syn,\beta}(\widetilde{\boldsymbol{x}}|D_{\tau-syn})}{1-\gamma^2}}$$

$$\leq \sqrt{\frac{M\Gamma_{syn}\mathbb{I}_{\tau-syn}(\widetilde{\boldsymbol{x}};f)}{1-\gamma^2}}+\frac{\epsilon}{1-\gamma}$$

Therefore, we have:

$$\mathbb{E}_{\tau-syn}\left[\sum_{t=\tau}^{\infty}\sum_{m=1}^{M}\gamma^{M(t-\tau)+m-1}\left(f(x_*) - f(x_{M(t-1)+m})\right)\right]$$

$$\leq 2\sqrt{\frac{M\Gamma_{syn}\mathbb{I}_{\tau-syn}(\widetilde{x}; f)}{1-\gamma^2}} + \frac{2\epsilon}{1-\gamma}$$

**Theorem 2.** (Undiscounted regret for the synchronous setting) If $\beta = \frac{T}{\Gamma_{syn}}$, for all target points $\widetilde{x}$ and $\tau = 1, 2, \cdots$, we have:

$$\mathbb{E}_{\tau-syn}\left[\sum_{t=\tau}^{T+\tau}\sum_{m=1}^{M} f(x_*) - f(x_{M(t-1)+m})\right] \leq 2\sqrt{M\Gamma_{syn}MT\mathbb{I}_{\tau-syn}(\widetilde{x}; f)} + 2MT\epsilon,$$

where $\epsilon = \sqrt{\mathbb{E}_{\tau-syn}\left[(f(x_*) - f(\widetilde{x}))^2\right]}$.

*Proof*: First, we have:

$$\mathbb{E}_{\tau-syn}\left[\sum_{t=\tau}^{T+\tau}\sum_{m=1}^{M} f(x_*) - f(x_{M(t-1)+m})\right]$$

$$= \mathbb{E}_{\tau-syn}\left[\sum_{t=\tau}^{T+\tau}\sum_{m=1}^{M}\left(f(x_*) - f(\widetilde{x}_{t-syn}) + f(\widetilde{x}_{t-syn}) - f(x_{M(t-1)+m})\right)\right]$$

$$= \mathbb{E}_{\tau-syn}\left[\sum_{t=\tau}^{T+\tau}\sum_{m=1}^{M}\left(f(x_*) - f(\widetilde{x}_{t-syn})\right)\right]$$

$$+ \mathbb{E}_{\tau-syn}\left[\sum_{t=\tau}^{T+\tau}\sum_{m=1}^{M}\left(f(\widetilde{x}_{t-syn}) - f(x_{M(t-1)+m})\right)\right]$$

Given the definition of $\mathcal{L}_{syn,\beta}(\widetilde{x}_{t-syn}|D_{t-syn})$ and the value of $\beta$, we have:

$$\mathcal{L}_{syn,\beta}(\widetilde{x}_{t-syn}|D_{t-syn}) = \mathbb{I}_{t-syn}(\widetilde{x}_{t-syn}; f) + \beta \mathbb{E}_{t-syn}\left[(f(x_*) - f(\widetilde{x}_{t-syn}))^2\right]$$

$$\geq \frac{T}{\Gamma_{syn}}\mathbb{E}_{t-syn}\left[(f(x_*) - f(\widetilde{x}_{t-syn}))^2\right]$$

Using Jensen's inequality and the fact that $\widetilde{x}_{t-syn}$ minimizes $\mathcal{L}_{syn,\beta}(\widetilde{x}_{t-syn}|D_{t-syn})$, we have:

$$\mathbb{E}_{t-syn}[f(x_*) - f(\widetilde{x}_{t-syn})] \leq \sqrt{\mathbb{E}_{t-syn}\left[(f(x_*) - f(\widetilde{x}_{t-syn}))^2\right]}$$

$$\leq \sqrt{\frac{\Gamma_{syn}}{T}\mathcal{L}_{syn,\beta}(\widetilde{x}_{t-syn}|D_{t-syn})}$$

$$\leq (MT)^{-1}\sqrt{M\Gamma_{syn}MT\mathcal{L}_{syn,\beta}(\widetilde{x}|D_{t-syn})}$$

Lemma 1 implies that for all $t \geq \tau$:

$$\mathbb{E}_{\tau-syn}[\mathcal{L}_{syn,\beta}(\widetilde{x}|D_{t-syn})] \leq \mathcal{L}_{syn,\beta}(\widetilde{x}|D_{\tau-syn}).$$

Combined with Jensen's inequality, we have:

$$\mathbb{E}_{\tau-syn}[f(x_*) - f(\widetilde{x}_{t-syn})] \leq (MT)^{-1}\mathbb{E}_{\tau-syn}\left[\sqrt{M\Gamma_{syn}MT\mathcal{L}_{syn,\beta}(\widetilde{x}|D_{t-syn})}\right]$$

$$\leq (MT)^{-1}\sqrt{M\Gamma_{syn}MT\mathbb{E}_{\tau-syn}[\mathcal{L}_{syn,\beta}(\widetilde{x}|D_{t-syn})]}$$

$$\leq (MT)^{-1}\sqrt{M\Gamma_{syn}MT\mathcal{L}_{syn,\beta}(\widetilde{x}|D_{\tau-syn})}$$

It follows that:

$$\mathbb{E}_{\tau-syn}\left[\sum_{t=\tau}^{T+\tau}\sum_{m=1}^{M}\left(f(x_*) - f(\widetilde{x}_{t-syn})\right)\right] \leq \sqrt{M\Gamma_{syn}MT\mathcal{L}_{syn,\beta}(\widetilde{x}|D_{\tau-syn})}$$

$$\leq \sqrt{M\Gamma_{syn}MT(\mathbb{I}_{\tau-syn}(\widetilde{x}; f) + \beta\epsilon^2)}$$

$$\leq \sqrt{M\Gamma_{syn}MT\mathbb{I}_{\tau-syn}(\widetilde{x}; f)} + MT\epsilon$$

In addition, from the definition of $\Gamma_{syn}$, Corollary 1, and the Cauchy-Schwartz inequality, we have:

$$\mathbb{E}_{\tau-syn}\left[\sum_{t=\tau}^{T+\tau}\sum_{m=1}^{M}\left(f(\widetilde{x}_{t-syn}) - f(x_{M(t-1)+m})\right)\right]$$

$$\leq \mathbb{E}_{\tau-syn}\left[\sum_{t=\tau}^{T+\tau}\sum_{m=1}^{M}\sqrt{\Gamma_{syn}\mathbb{I}_{t-syn}\left(\widetilde{x}_{t-syn}; (X_t, Y_t)\right)}\right]$$

$$\leq \sqrt{\Gamma_{syn}MT\sum_{t=\tau}^{T+\tau}\sum_{m=1}^{M}\mathbb{E}_{\tau-syn}\left[\mathbb{I}_{t-syn}\left(\widetilde{x}_{t-syn}; (X_t, Y_t)\right)\right]}$$

$$\leq \sqrt{M\Gamma_{syn}MT\mathcal{L}_{syn,\beta}(\widetilde{x}|D_{\tau-syn})}$$

$$\leq \sqrt{M\Gamma_{syn}MT\mathbb{I}_{\tau-syn}(\widetilde{x}; f)} + MT\epsilon$$

Therefore, we have:

$$\mathbb{E}_{\tau-syn}\left[\sum_{t=\tau}^{T+\tau}\sum_{m=1}^{M} f(x_*) - f(x_{M(t-1)+m})\right] \leq 2\sqrt{M\Gamma_{syn}MT\mathbb{I}_{\tau-syn}(\tilde{x}; f)} + 2MT\epsilon$$

*A.2. Proofs of regret bounds for the asynchronous version of STS-PBO*

**Lemma 3.** For all $\beta > 0$, target points $\tilde{x}$ and $t = M, M+1, \cdots$, we have:

$$\mathbb{E}_{t-asy}[\mathcal{L}_{asy,\beta}(\tilde{x}|D_{(t+1)-asy})] = \mathcal{L}_{asy,\beta}(\tilde{x}|D_{t-asy}) - \mathbb{I}_{t-asy}\left(\tilde{x}; (x'_{t-asy}, y'_{t-asy})\right).$$

*Proof:* Recall that $D_{(t+1)-asy} = (D_{t-asy}, x'_{t-asy}, y'_{t-asy})$. Here the subscript $t - asy$ stands for $t - M + 1$ in the evaluation results. As mentioned in section 3.1, we have $\tilde{x} \perp D_t | f, \forall t$. Therefore, we have:

$$\mathbb{I}_{t-asy}(\tilde{x}; (x'_{t-asy}, y'_{t-asy}) | f) = 0.$$

By using the chain rule of mutual information, we have:

$$\mathbb{I}_{t-asy}(\tilde{x}; f) = \mathbb{I}_{t-asy}(\tilde{x}; f) + \mathbb{I}_{t-asy}(\tilde{x}; (x'_{t-asy}, y'_{t-asy}) | f)$$

$$= \mathbb{I}_{t-asy}\left(\tilde{x}; f, (x'_{t-asy}, y'_{t-asy})\right)$$

$$= \mathbb{I}_{t-asy}\left(\tilde{x}; f | (x'_{t-asy}, y'_{t-asy})\right) + \mathbb{I}_{t-asy}\left(\tilde{x}; (x'_{t-asy}, y'_{t-asy})\right)$$

It follows that:

$$\mathbb{E}_{t-asy}[\mathcal{L}_{asy,\beta}(\tilde{x}|D_{(t+1)-asy})]$$

$$= \mathbb{E}[\mathcal{L}_{asy,\beta}(\tilde{x}|D_{(t+1)-asy})|D_{t-asy}]$$

$$= \mathbb{E}\left[\mathbb{I}_{t-asy}\left(\tilde{x}; f | (x'_{t-asy}, y'_{t-asy})\right) + \beta\mathbb{E}\left[(f(x_*) - f(\tilde{x}))^2 | D_{(t+1)-asy}\right] | D_{t-asy}\right]$$

$$= \mathbb{E}_{t-asy}\left[\mathbb{I}_{t-asy}\left(\tilde{x}; f | (x'_{t-asy}, y'_{t-asy})\right)\right] + \beta\mathbb{E}_{t-asy}\left[(f(x_*) - f(\tilde{x}))^2\right]$$

$$= \mathbb{E}_{t-asy}\left[\mathbb{I}_{t-asy}(\tilde{x}; f) - \mathbb{I}_{t-asy}\left(\tilde{x}; (x'_{t-asy}, y'_{t-asy})\right)\right]$$

$$\quad + \beta\mathbb{E}_{t-asy}\left[(f(x_*) - f(\tilde{x}))^2\right]$$

$$= \mathbb{I}_{t-asy}(\tilde{x}; f) + \beta\mathbb{E}_{t-asy}\left[(f(x_*) - f(\tilde{x}))^2\right] - \mathbb{I}_{t-asy}\left(\tilde{x}; (x'_{t-asy}, y'_{t-asy})\right)$$

$$= \mathcal{L}_{asy,\beta}(\tilde{x}|D_{t-asy}) - \mathbb{I}_{t-asy}\left(\tilde{x}; (x'_{t-asy}, y'_{t-asy})\right)$$

**Lemma 4.** For all $\beta > 0$, target points $\tilde{x}$ and $t = M, M+1, \cdots$, we have:

$$\mathbb{E}_{t-asy}[\mathcal{L}_{asy,\beta}(\tilde{x}_{t+1}|D_{(t+1)-asy})] \leq \mathcal{L}_{asy,\beta}(\tilde{x}_t|D_{t-asy}) -$$

$$\mathbb{I}_{t-asy}\left(\tilde{x}_t; (x'_{t-asy}, y'_{t-asy})\right).$$

*Proof*: Recall that by definition $\tilde{x}_{t+1}$ minimizes $\mathcal{L}_{asy,\beta}(\tilde{x}_{t+1}|D_{(t+1)-asy})$, we then have:

$$\mathbb{E}_{t-asy}[\mathcal{L}_{asy,\beta}(\tilde{x}_{t+1}|D_{(t+1)-asy})] \leq \mathbb{E}[\mathcal{L}_{asy,\beta}(\tilde{x}_t|D_{(t+1)-asy})|D_{t-asy}] =$$

$$\mathcal{L}_{asy,\beta}(\tilde{x}_t|D_{t-asy}) - \mathbb{I}_{t-asy}\left(\tilde{x}_t; (x'_{t-asy}, y'_{t-asy})\right).$$

**Corollary 2.** For all $\beta > 0$, target points $\tilde{x}$ and $\tau = M, M+1, \cdots$, we have:

$$\mathbb{E}_{\tau-asy}\left[\sum_{t=\tau}^{\infty} \mathbb{I}_{t-asy}\left(\tilde{x}_t; (x'_{t-asy}, y'_{t-asy})\right)\right] \leq \mathcal{L}_{asy,\beta}(\tilde{x}|D_{\tau-asy}),$$

$$\mathbb{E}_{\tau-asy}\left[\sum_{t=\tau}^{T+\tau} \mathbb{I}_{t-asy}\left(\tilde{x}_t; (x'_{t-asy}, y'_{t-asy})\right)\right] \leq \mathcal{L}_{asy,\beta}(x|D_{\tau-asy}).$$

*Proof*: $\mathbb{E}_{\tau-asy}\left[\sum_{t=\tau}^{\infty} \mathbb{I}_{t-asy}\left(\tilde{x}_t; (x'_{t-asy}, y'_{t-asy})\right)\right]$

$$\leq \mathbb{E}_{\tau-asy}\left[\sum_{t=\tau}^{\infty} \mathcal{L}_{asy,\beta}(\tilde{x}_t|D_{t-asy}) - \mathbb{E}_{t-asy}[\mathcal{L}_{asy,\beta}(\tilde{x}_{t+1}|D_{(t+1)-asy})]\right]$$

$$= \sum_{t=\tau}^{\infty} \mathbb{E}_{\tau-asy}[\mathcal{L}_{asy,\beta}(\tilde{x}_t|D_{t-asy})] - \mathbb{E}_{\tau-asy}\left[\mathbb{E}_{t-asy}[\mathcal{L}_{asy,\beta}(\tilde{x}_{t+1}|D_{(t+1)-asy})]\right]$$

$$= \mathbb{E}_{\tau-asy}[\mathcal{L}_{asy,\beta}(\tilde{x}_\tau|D_{\tau-asy})] + \sum_{t=\tau+1}^{\infty} \mathbb{E}_{\tau-asy}[\mathcal{L}_{asy,\beta}(\tilde{x}_t|D_{t-asy})]$$

$$- \sum_{t=\tau}^{\infty} \mathbb{E}_{\tau-asy}[\mathcal{L}_{asy,\beta}(\tilde{x}_{t+1}|D_{(t+1)-asy})]$$

$$= \mathcal{L}_{asy,\beta}(\tilde{x}_\tau|D_{\tau-asy})$$

$$\leq \mathcal{L}_{asy,\beta}(\tilde{x}|D_{\tau-asy})$$

where the proofs follow the fact that $\tilde{x}_\tau$ minimizes $\mathcal{L}_{asy,\beta}(\tilde{x}_\tau|D_{\tau-asy})$. Then we have:

$$\mathbb{E}_{\tau-asy}\left[\sum_{t=\tau}^{T+\tau} \mathbb{I}_{t-asy}\left(\tilde{x}_t; (x'_{t-asy}, y'_{t-asy})\right)\right]$$

$$\leq \mathbb{E}_{\tau-asy}\left[\sum_{t=\tau}^{\infty} \mathbb{I}_{t-asy}\left(\tilde{x}_t; (x'_{t-asy}, y'_{t-asy})\right)\right]$$

$$\leq \mathcal{L}_{asy,\beta}(\tilde{x}|D_{\tau-asy})$$

**Theorem 3.** (Discounted regret for the asynchronous setting) If $\beta = \frac{1-\gamma^2}{(1-\gamma)^2 \Gamma_{asy}}$, for all target points $\tilde{x}$ and $\tau = M, M+1, \cdots$, we have:

$$\mathbb{E}_{\tau-asy}\left[\sum_{t=\tau}^{\infty} \gamma^{t-\tau}(f(x_*) - f(x_t))\right] \leq 2\sqrt{\frac{\Gamma_{asy}\mathbb{I}_{\tau-asy}(x;f)}{1-\gamma^2}} + \frac{2\epsilon}{1-\gamma},$$

where $\epsilon = \sqrt{\mathbb{E}_{\tau-asy}\left[(f(x_*) - f(\tilde{x}))^2\right]}$.

*Proof*: First, we have:

$$\mathbb{E}_{\tau-asy}\left[\sum_{t=\tau}^{\infty} \gamma^{t-\tau}(f(x_*) - f(x_t))\right]$$

$$= \mathbb{E}_{\tau-asy}\left[\sum_{t=\tau}^{\infty} \gamma^{t-\tau}(f(x_*) - f(\tilde{x}_t) + f(\tilde{x}_t) - f(x_t))\right]$$

$$= \mathbb{E}_{\tau-asy}\left[\sum_{t=\tau}^{\infty} \gamma^{t-\tau}(f(x_*) - f(\tilde{x}_t))\right] + \mathbb{E}_{\tau-asy}\left[\sum_{t=\tau}^{\infty} \gamma^{t-\tau}(f(\tilde{x}_t) - f(x_t))\right]$$

Given the definition of $\mathcal{L}_{asy,\beta}(\tilde{x}_t|D_{t-asy})$ and the value of $\beta$, we have:

$$\mathcal{L}_{asy,\beta}(\tilde{x}_t|D_{t-asy}) = \mathbb{I}_{t-asy}(\tilde{x}_t;f) + \beta \mathbb{E}_{t-asy}\left[(f(x_*) - f(\tilde{x}_t))^2\right]$$

$$\geq \frac{1-\gamma^2}{(1-\gamma)^2 \Gamma_{asy}} \mathbb{E}_{t-asy}\left[(f(x_*) - f(\tilde{x}_t))^2\right]$$

Using Jensen's inequality and the fact that $\tilde{x}_t$ minimizes $\mathcal{L}_{asy,\beta}(\tilde{x}_t|D_{t-asy})$, we have:

$$\mathbb{E}_{t-asy}[f(x_*) - f(\tilde{x}_t)] \leq \sqrt{\mathbb{E}_{t-asy}\left[(f(x_*) - f(\tilde{x}_t))^2\right]}$$

$$\leq \sqrt{\frac{(1-\gamma)^2 \Gamma_{asy}}{1-\gamma^2} \mathcal{L}_{asy,\beta}(\tilde{x}_t|D_{t-asy})}$$

$$\leq (1-\gamma)\sqrt{\frac{\Gamma_{asy}\mathcal{L}_{asy,\beta}(\tilde{x}|D_{t-asy})}{1-\gamma^2}}$$

Lemma 3 implies that for all $t \geq \tau$:

$$\mathbb{E}_{\tau-asy}[\mathcal{L}_{asy,\beta}(\tilde{x}|D_{t-asy})] \leq \mathcal{L}_{asy,\beta}(\tilde{x}|D_{\tau-asy}).$$

Combined with Jensen's inequality, we have:

$$\mathbb{E}_{\tau-asy}[f(\boldsymbol{x}_*) - f(\tilde{\boldsymbol{x}}_t)] \leq (1-\gamma)\mathbb{E}_{\tau-asy}\left[\sqrt{\frac{\Gamma_{asy}\mathcal{L}_{asy,\beta}(\tilde{\boldsymbol{x}}|D_{t-asy})}{1-\gamma^2}}\right]$$

$$\leq (1-\gamma)\sqrt{\frac{\Gamma_{asy}\mathbb{E}_{\tau-asy}[\mathcal{L}_{asy,\beta}(\tilde{\boldsymbol{x}}|D_{t-asy})]}{1-\gamma^2}}$$

$$\leq (1-\gamma)\sqrt{\frac{\Gamma_{asy}\mathcal{L}_{asy,\beta}(\tilde{\boldsymbol{x}}|D_{\tau-asy})}{1-\gamma^2}}$$

It follows that:

$$\mathbb{E}_{\tau-asy}\left[\sum_{t=\tau}^{\infty}\gamma^{t-\tau}(f(\boldsymbol{x}_*) - f(\tilde{\boldsymbol{x}}_t))\right]$$

$$\leq \sqrt{\frac{\Gamma_{asy}\mathcal{L}_{asy,\beta}(\tilde{\boldsymbol{x}}|D_{\tau-asy})}{1-\gamma^2}}$$

$$\leq \sqrt{\frac{\Gamma_{asy}(\mathbb{I}_{\tau-asy}(\tilde{\boldsymbol{x}}; f) + \beta\epsilon^2)}{1-\gamma^2}}$$

$$\leq \sqrt{\frac{\Gamma_{asy}\mathbb{I}_{\tau-asy}(\tilde{\boldsymbol{x}}; f)}{1-\gamma^2}} + \frac{\epsilon}{1-\gamma}$$

In addition, from the definition of $\Gamma_{asy}$, Corollary 2, and the Cauchy-Schwartz inequality, we have:

$$\mathbb{E}_{\tau-asy}\left[\sum_{t=\tau}^{\infty}\gamma^{t-\tau}(f(\tilde{\boldsymbol{x}}_t) - f(\boldsymbol{x}_t))\right]$$

$$\leq \mathbb{E}_{\tau-asy}\left[\sum_{t=\tau}^{\infty}\gamma^{t-\tau}\sqrt{\Gamma_{asy}\mathbb{I}_{t-asy}(\tilde{\boldsymbol{x}}_t; (\boldsymbol{x}'_{t-asy}, y'_{t-asy}))}\right]$$

$$\leq \sqrt{\sum_{t=\tau}^{\infty}\gamma^{2(t-\tau)}\Gamma_{asy}\sum_{t=\tau}^{\infty}\mathbb{E}_{\tau-asy}\left[\mathbb{I}_{t-asy}(\tilde{\boldsymbol{x}}_t; (\boldsymbol{x}'_{t-asy}, y'_{t-asy}))\right]}$$

$$\leq \sqrt{\Gamma_{asy}\mathcal{L}_{asy,\beta}(\tilde{\boldsymbol{x}}|D_{\tau-asy})\sum_{t=0}^{\infty}\gamma^{2t}}$$

$$= \sqrt{\frac{\Gamma_{asy}\mathcal{L}_{asy,\beta}(\tilde{x}|D_{\tau-asy})}{1-\gamma^2}}$$

$$\leq \sqrt{\frac{\Gamma_{asy}\mathbb{I}_{\tau-asy}(\tilde{x}; f)}{1-\gamma^2}} + \frac{\epsilon}{1-\gamma}$$

Therefore, we have:

$$\mathbb{E}_{\tau-asy}\left[\sum_{t=\tau}^{\infty}\gamma^{t-\tau}(f(x_*) - f(x_t))\right] \leq 2\sqrt{\frac{\Gamma_{asy}\mathbb{I}_{\tau-asy}(\tilde{x}; f)}{1-\gamma^2}} + \frac{2\epsilon}{1-\gamma}$$

**Theorem 4.** (Undiscounted regret for the asynchronous setting) If $\beta = \frac{T}{\Gamma_{asy}}$, for all target points $\tilde{x}$ and $\tau = M, M+1, \cdots$, we have:

$$\mathbb{E}_{\tau-asy}[\sum_{t=\tau}^{T+\tau} f(x_*) - f(x_t)] \leq 2\sqrt{\Gamma_{asy}T\mathbb{I}_{\tau-asy}(x; f)} + 2T\epsilon,$$

where $\epsilon = \sqrt{\mathbb{E}_{\tau-asy}\left[(f(x_*) - f(\tilde{x}))^2\right]}$.

*Proof*: First, we have

$$\mathbb{E}_{\tau-asy}\left[\sum_{t=\tau}^{T+\tau}(f(x_*) - f(x_t))\right] = \mathbb{E}_{\tau-asy}\left[\sum_{t=\tau}^{T+\tau}(f(x_*) - f(\tilde{x}_t) + f(\tilde{x}_t) - f(x_t))\right]$$

$$= \mathbb{E}_{\tau-asy}\left[\sum_{t=\tau}^{T+\tau}(f(x_*) - f(\tilde{x}_t))\right]$$

$$+ \mathbb{E}_{\tau-asy}\left[\sum_{t=\tau}^{T+\tau}(f(\tilde{x}_t) - f(x_t))\right]$$

Given the definition of $\mathcal{L}_{asy,\beta}(\tilde{x}_t|D_{t-asy})$ and the value of $\beta$, we have:

$$\mathcal{L}_{asy,\beta}(\tilde{x}_t|D_{t-asy}) = \mathbb{I}_{t-asy}(\tilde{x}_t; f) + \beta\mathbb{E}_{t-asy}\left[(f(x_*) - f(\tilde{x}_t))^2\right]$$

$$\geq \frac{T}{\Gamma_{asy}}\mathbb{E}_{t-asy}\left[(f(x_*) - f(\tilde{x}_t))^2\right]$$

Using Jensen's inequality and the fact that $\tilde{x}_t$ minimizes $\mathcal{L}_{asy,\beta}(\tilde{x}_t|D_{t-asy})$, we have:

$$\mathbb{E}_{t-asy}[f(x_*) - f(\tilde{x}_t)]$$

$$\leq \sqrt{\mathbb{E}_{t-asy}\left[(f(x_*) - f(\tilde{x}_t))^2\right]}$$

$$\leq \sqrt{\frac{\Gamma_{asy}}{T}\mathcal{L}_{asy,\beta}(\widetilde{\pmb{x}}_t|D_{t-asy})}$$

$$\leq \sqrt{\frac{\Gamma_{asy}}{T}\mathcal{L}_{asy,\beta}(\widetilde{\pmb{x}}|D_{t-asy})}$$

Lemma 3 implies that for all $t \geq \tau$:

$$\mathbb{E}_{\tau-asy}[\mathcal{L}_{asy,\beta}(\widetilde{\pmb{x}}|D_{t-asy})] \leq \mathcal{L}_{asy,\beta}(\widetilde{\pmb{x}}|D_{\tau-asy}).$$

Combined with Jensen's inequality, we have:

$$\mathbb{E}_{\tau-asy}[f(\pmb{x}_*) - f(\widetilde{\pmb{x}}_t)] \leq T^{-1}\mathbb{E}_{\tau-asy}\left[\sqrt{\Gamma_{asy}T\mathcal{L}_{asy,\beta}(\widetilde{\pmb{x}}|D_{t-asy})}\right]$$

$$\leq T^{-1}\sqrt{\Gamma_{asy}T\mathbb{E}_{\tau-asy}[\mathcal{L}_{asy,\beta}(\widetilde{\pmb{x}}|D_{t-asy})]}$$

$$\leq T^{-1}\sqrt{\Gamma_{asy}T\mathcal{L}_{asy,\beta}(\widetilde{\pmb{x}}|D_{\tau-asy})}$$

It follows that:

$$\mathbb{E}_{\tau-asy}\left[\sum_{t=\tau}^{T+\tau}(f(\pmb{x}_*) - f(\widetilde{\pmb{x}}_t))\right] \leq \sqrt{\Gamma_{asy}T\mathcal{L}_{asy,\beta}(\widetilde{\pmb{x}}|D_{\tau-asy})}$$

$$\leq \sqrt{\Gamma_{asy}T\left(\mathbb{I}_{\tau-asy}(\widetilde{\pmb{x}};f) + \beta\epsilon^2\right)}$$

$$\leq \sqrt{\Gamma_{asy}T\mathbb{I}_{\tau-asy}(\widetilde{\pmb{x}};f)} + T\epsilon$$

In addition, from the definition of $\Gamma_{asy}$, Corollary 2, and the Cauchy-Schwartz inequality, we have:

$$\mathbb{E}_{\tau-asy}\left[\sum_{t=\tau}^{T+\tau}f(\pmb{x}_*) - f(\pmb{x}_t)\right] \leq \mathbb{E}_{\tau-asy}\left[\sum_{t=\tau}^{T+\tau}\sqrt{\Gamma_{asy}\mathbb{I}_{t-asy}\left(\widetilde{\pmb{x}}_t;(\pmb{x}'_{t-asy},y'_{t-asy})\right)}\right]$$

$$\leq \sqrt{\Gamma_{asy}T\sum_{t=\tau}^{T+\tau}\mathbb{E}_{\tau-asy}\left[\mathbb{I}_{t-asy}\left(\widetilde{\pmb{x}}_t;(\pmb{x}'_{t-asy},y'_{t-asy})\right)\right]}$$

$$\leq \sqrt{\Gamma_{asy}T\mathcal{L}_{asy,\beta}(\widetilde{\pmb{x}}|D_{\tau-asy})}$$

$$\leq \sqrt{\Gamma_{asy}T\mathbb{I}_{\tau-asy}(\widetilde{\pmb{x}};f)} + T\epsilon$$

Therefore, we have:

$$\mathbb{E}_{\tau-asy}\left[\sum_{t=\tau}^{T+\tau}(f(\pmb{x}_*) - f(\pmb{x}_t))\right] \leq 2\sqrt{\Gamma_{asy}T\mathbb{I}_{\tau-asy}(\widetilde{\pmb{x}}; f)} + 2T\epsilon$$

**Acknowledgments**

This work was supported by the National Natural Science Foundation of China under grant number 62273197, the National Key Research and Development Program of China under grant number 2022YFE0197600, and the Beijing National Research Center for Information Science and Technology under grant number BNR2023RC01008.

**References:**

[1] Snoek, J., Larochelle, H., & Adams, R. P. (2012). Practical Bayesian optimization of machine learning algorithms. Advances in Neural Information Processing Systems, 25.

[2] Jiang, B., Berliner, M. D., Lai, K., Asinger, P. A., Zhao, H., Herring, P. K., et al. (2022). Fast charging design for Lithium-ion batteries via Bayesian optimization. Applied Energy, 307, 118244.

[3] Olofsson, S., Mehrian, M., Calandra, R., Geris, L., Deisenroth, M. P., & Misener, R. (2018). Bayesian multiobjective optimisation with mixed analytical and black-box functions: Application to tissue engineering. IEEE Transactions on Biomedical Engineering, 66(3), 727–739.

[4] Martinez-Cantin, R., De Freitas, N., Brochu, E., Castellanos, J., & Doucet, A. (2009). A Bayesian exploration-exploitation approach for optimal online sensing and planning with a visually guided mobile robot. Autonomous Robots, 27, 93–103.

[5] Chakrabarty, A., & Benosman, M. (2021). Safe learning-based observers for unknown nonlinear systems using Bayesian optimization. Automatica, 133, 109860.

[6] Neumann-Brosig, M., Marco, A., Schwarzmann, D., & Trimpe, S. (2019). Data-efficient autotuning with Bayesian optimization: An industrial control study. IEEE Transactions on Control Systems Technology, 28(3), 730–740.

[7] Arumugam, D., & Van Roy, B. (2021). Deciding what to learn: A rate-distortion approach. In International Conference on Machine Learning, 373–382.

[8] Jiang, B., Gent, W. E., Mohr, F., Das, S., Berliner, M. D., Forsuelo, M., et al. (2021). Bayesian learning for rapid prediction of lithium-ion battery-cycling protocols. Joule, 5(12), 3187–3203.

[9] Russo, D., & Van Roy, B. (2018). Satisficing in time-sensitive bandit learning. arXiv preprint arXiv:1803.02855.

[10] Thompson, W. R. (1933). On the likelihood that one unknown probability exceeds another in view of the evidence of two samples. Biometrika, 25(3), 285–294.

[11] Russo, D., & Van Roy, B. (2016). An information-theoretic analysis of Thompson sampling. The Journal of Machine Learning Research, 17(1), 2442–2471.

[12] Shannon, C. E. (1959). Coding theorems for a discrete source with a fidelity criterion. In IRE International Convention Record, 4, 142–163.

[13] Berger, T. (1971). Rate Distortion Theory: A Mathematical Basis for Data Compression. Prentice-Hall.

[14] Blahut, R. (1972). Computation of channel capacity and rate-distortion functions. IEEE Transactions on Information Theory, 18(4), 460–473.

[15] Arimoto, S. (1972). An algorithm for computing the capacity of arbitrary discrete memoryless channels. IEEE Transactions on Information Theory, 18(1), 14–20.

[16] Arumugam, D., & Van Roy, B. (2021). The value of information when deciding what to learn. Advances in Neural Information Processing Systems, 34, 9816–9827.

[17] Russo, D., & Van Roy, B. (2018). Learning to optimize via information-directed sampling. Operations Research, 66(1), 230–252.

[18] Kandasamy, K., Krishnamurthy, A., Schneider, J., & Póczos, B. (2018). Parallelised Bayesian optimisation via Thompson sampling. In International Conference on Artificial Intelligence and Statistics, 133–142.

[19] Ginsbourger, D., Janusevskis, J., & Le Riche, R. (2011). Dealing with asynchronicity in parallel Gaussian process based global optimization. Doctoral Dissertation, Mines Saint-Etienne.

[20] Contal, E., Buffoni, D., Robicquet, A., & Vayatis, N. (2013). Parallel Gaussian

process optimization with upper confidence bound and pure exploration. In Machine Learning and Knowledge Discovery in Databases: European Conference, 225–240.

[21] Wang, Z., Gehring, C., Kohli, P., & Jegelka, S. (2018). Batched large-scale Bayesian optimization in high-dimensional spaces. In International Conference on Artificial Intelligence and Statistics, 745–754.

[22] Wang, J., Clark, S. C., Liu, E., & Frazier, P. I. (2020). Parallel Bayesian global optimization of expensive functions. Operations Research, 68(6), 1850–1865.

[23] Satake, M., Takahashi, K., Shimomura, Y., & Takizawa, H. (2023). Balancing exploitation and exploration in parallel Bayesian optimization under computing resource constraint. In 2023 IEEE International Parallel and Distributed Processing Symposium Workshops, 706–713.

[24] Jones, D. R., Schonlau, M., & Welch, W. J. (1998). Efficient global optimization of expensive black-box functions. Journal of Global optimization, 13, 455–492.

[25] Srinivas, N., Krause, A., Kakade, S., & Seeger, M. (2010). Gaussian process optimization in the bandit setting: no regret and experimental design. In Proceedings of the 27th International Conference on International Conference on Machine Learning, 1015–1022.

[26] Rasmussen, C. E., & Williams, C. K. (2006). Gaussian Processes for Machine Learning. Cambridge, MA: MIT Press.

[27] Russo, D., & Van Roy, B. (2014). Learning to optimize via posterior sampling. Mathematics of Operations Research, 39(4), 1221–1243.

[28] Li, H., Zhang, X., Peng, J., He, J., Huang, Z., & Wang, J. (2020). Cooperative CC–CV charging of supercapacitors using multicharger systems. IEEE Transactions on Industrial Electronics, 67(12), 10497–10508.

[29] Berliner, M. D., Cogswell, D. A., Bazant, M. Z., & Braatz, R. D. (2021). Methods—PETLION: Open-source software for millisecond-scale porous electrode theory-based lithium-ion battery simulations. Journal of The Electrochemical Society, 168(9), 090504.

[30] Jiang, B., Wang, Y., Ma, Z., & Lu, Q. (2023). Fast charging of lithium-ion batteries using deep Bayesian optimization with recurrent neural network. arXiv preprint

arXiv:2304.04195.

[31] Li, Z., Dong, Z., Liang, Z., & Ding, Z. (2021). Surrogate-based distributed optimisation for expensive black-box functions. Automatica, 125, 109407.

[32] Wang, Z., Lai, X., & Wu, Q. (2016). A PSR CC/CV flyback converter with accurate CC control and optimized CV regulation strategy. IEEE Transactions on Power Electronics, 32(9), 7045–7055.